\documentclass[10pt,twocolumn,letterpaper]{article}

\usepackage{cvpr}              

\def\confName{CVPR}

\definecolor{cvprblue}{rgb}{0.21,0.49,0.74}
\usepackage[pagebackref,breaklinks,colorlinks,allcolors=cvprblue]{hyperref}

\usepackage{multirow}
\usepackage{graphicx}
\usepackage[accsupp]{axessibility}  

\usepackage{rotating}
\usepackage{pifont}
\usepackage[acronym, shortcuts]{glossaries}
\usepackage{tikz}
\usetikzlibrary{trees}
\RequirePackage{siunitx} 

\newcommand{\cmark}{\textcolor{green!60!black}{\ding{51}}}  
\newcommand{\xmark}{\textcolor{red!70!black}{\ding{55}}}

\newacronym{eo}{EO}{Earth Observation}
\newacronym{rs}{RS}{Remote Sensing}
\newacronym{dl}{DL}{Deep Learning}
\newacronym{ml}{ML}{Machine Learning}
\newacronym{de}{DE}{Deep Ensemble}
\newacronym{s2}{S2}{Sentinel-2}
\newacronym{adm}{ADM}{Additional Data Modalitie}

\def\paperID{Anonymous} 
\def\confName{CVPR}
\def\confYear{2026}

\title{
YieldSAT: A Multimodal Benchmark Dataset for High-Resolution Crop Yield Prediction
}

\author{
Miro Miranda$^{1,2,}$\thanks{corresponding authors: {\tt\small\{ miro.miranda\_lorenz, deepak\_kumar.pathak\}@dfki.de}  } , 
Deepak Pathak$^{2,*}$, 
Patrick Helber$^{3}$, 
Benjamin Bischke$^{3}$,  
Hiba Najjar$^{1,2}$, \\
Francisco Mena$^{1,2}$, 
Cristhian Sanchez$^{2}$, 
Akshay Pai$^{2}$, 
Diego Arenas$^{2}$, \\
Matias Valdenegro-Toro$^{4}$, 
Marcela Charfuelan$^{2}$, 
Marlon Nuske$^{2}$, 
Andreas Dengel$^{1,2}$ \\
$^{1}$RPTU Kaiserslautern-Landau \quad
$^{2}$DFKI GmbH \quad
$^{3}$Vision Impulse GmbH \quad
$^{4}$University of Groningen 
}

\begin{document}
\maketitle
\begin{abstract}
Crop yield prediction requires substantial data to train scalable models. However, creating yield prediction datasets is constrained by high acquisition costs, heterogeneous data quality, and data privacy regulations. Consequently, existing datasets are scarce, low in quality, or limited to regional levels or single crop types, hindering the development of scalable data-driven solutions. 
In this work, we release \textit{YieldSAT}, a large, high-quality, and multimodal dataset for high-resolution crop yield prediction. \textit{YieldSAT} spans various climate zones across multiple countries, including Argentina, Brazil, Uruguay, and Germany, and includes major crop types, including corn, rapeseed, soybeans, and wheat, across 2,173 expert-curated fields. In total, over 12.2 million yield samples are available, each with a spatial resolution of $\SI{10}{m}$. 
Each field is paired with multispectral satellite imagery, resulting in 113,555 labeled satellite images, complemented by auxiliary environmental data. 
We demonstrate the potential of large-scale and high-resolution crop yield prediction as a pixel regression task by comparing various deep learning models and data fusion architectures. Furthermore, we highlight open challenges arising from severe distribution shifts in the ground truth data under real-world conditions. To mitigate this, we explore a domain-informed Deep Ensemble approach that exhibits significant performance gains. The dataset is available at \url{https://yieldsat.github.io/}. 
\end{abstract}    
\section{Introduction}
\begin{figure*}[!t]
    \centering
    \includegraphics[width=.8\linewidth]{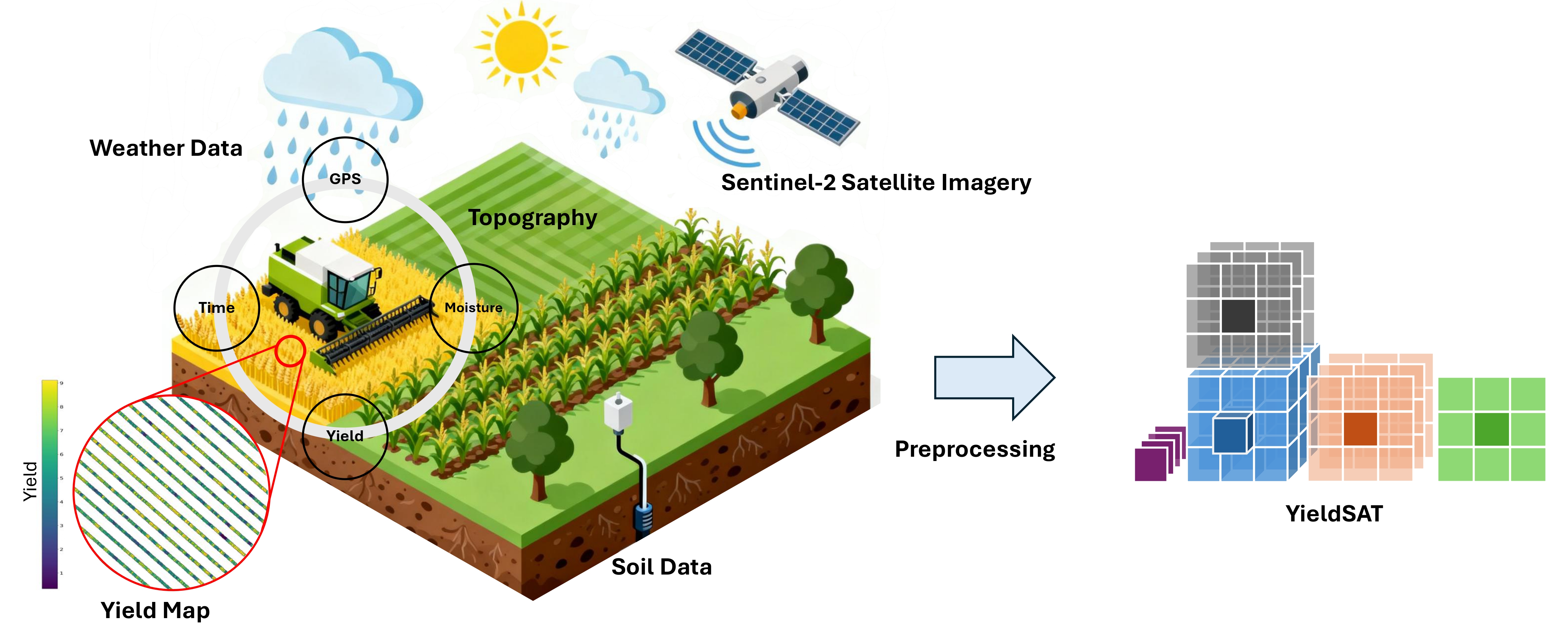}
    \caption{Schematic overview of the \textit{YieldSAT} dataset and the data collection and preprocessing. At harvest, a combine harvester collects point data containing various information (yield, geolocation, time, and moisture), referred to as a yield map. For each yield map, \gls{eo} data is collected (satellite imagery, weather, soil, and topography data). The target yield map and the input data are preprocessed into an ML-ready data format, formulating yield prediction as a pixel-wise regression task.}
    \label{fig:Yieldmap_with_data}
\end{figure*}
Digital agriculture has emerged as an essential tool for addressing current challenges in the agricultural sector, providing data-driven solutions for informed decision-making and ultimately for achieving the UN's Sustainable Development Goals (SDGs) \cite{roscher2023data}, specifically SDG 2 (no hunger) and SDG 13 (climate action) \cite{roscher2023data}. A key component is crop yield prediction at large scale and high spatial resolution, supporting the management and optimization of crop productivity, the implementation of regional policies, insurance concepts, and the adaptation to changing climate conditions \cite{miranda_drought}. Yield prediction can be considered as an image regression task involving the processing of multimodal time series data.
Nevertheless, yield prediction requires large amounts of high-quality data to train data-driven models \cite{mena2024adaptive}. 
In this context, the exponential growth of openly available \ac{rs} and \acrfull{eo} data has become an essential driver of recent advances in crop yield prediction. For instance, satellite programs like the \textit{\gls{s2} mission} of the \textit{Copernicus Program}
continuously deliver imagery in high spatial resolution and high temporal frequency  \cite{Copernicus_Annual_Senti}. Jointly, satellite data thoroughly captures crop development from seeding to harvesting by delivering information on soil properties, vegetation, water content, nutrient supply, and plant biochemistry
\cite{soille2018versatile,xie2019retrieval,metzger2021crop,crawford202350,hersbach2020era5}. \\
Although the amount of openly accessible \gls{eo} data is unlimited, labeled \gls{eo} datasets are highly scarce, making up only 0.1\% of the total volume of unlabeled data \cite{zhu2024foundations}, a reason why regression models remain largely underexplored \cite{xue2025regression}. In crop yield prediction, the lack of dedicated datasets is a serious concern, leading to models trained on single crop types, limited geographic locations, and individual years. Such models frequently exhibit severe performance collapse in real-world scenarios, leading to skepticism about their deployment in practice \cite{perich2023pixel,miranda_drought}. \\
To fill this gap, we created \textit{YieldSAT}, a high-quality dataset for large-scale, high-resolution crop yield prediction spanning 4 countries, 4 crop types, and 9 years. \textit{YieldSAT} is the first multimodal dataset for crop yield prediction at both field and subfield levels (i.e., pixel with \SI{10}{m} resolution), designed for supervised learning using only globally and publicly available input data. We provide an in-depth comparison of benchmark results obtained from multiple \ac{dl} architectures and data fusion methods. Moreover, we highlight open challenges in crop yield prediction arising from severe distribution shifts in the ground-truth data, which lead to significant performance collapse in \ac{dl} models. 
To find explanations and potential mitigation strategies, we propose a domain-informed \textit{\gls{de}} approach \cite{lakshminarayanan2017simple} and investigate the weight space distribution of the ensemble members.
Our approach demonstrates significant improvement relative to the baseline.\\
By releasing \textit{YieldSAT}, we aim to accelerate digital farming and \gls{eo} research. We believe this work supports the computer vision community in developing robust image regression methods that produce physically meaningful outputs under challenging, real-world conditions.

\section{Related Work}
Subfield (i.e., pixel) level crop yield prediction can be considered as a dense, structured image (pixel) regression task, similar to monocular depth estimation \cite{ke2024repurposing,eigen2014depth}. Nevertheless, yield prediction requires processing multimodal data, including long time series with varying temporal and spatial resolutions, making it a multimodal fusion task \cite{mena2024common}. Additionally, yield prediction is a physically grounded task that involves predicting biophysical quantities, connecting it to scientific methods like \textit{Physics-Informed Neural Networks (PINNs)} \cite{cuomo2022scientific,raissi2017physics,miranda2024exploring,miranda_drought} and explainable AI \cite{hohl2024opening,najjar2025explainability}. However, many \gls{eo} regression applications, like high-resolution yield prediction, remain underexplored, primarily due to the lack of dedicated datasets \cite{xue2025regression}. 
Still, yield prediction using \gls{dl} and \gls{eo} has gained widespread interest \cite{van2020crop}. 
For this, many methods for yield prediction use multispectral satellite imagery and advanced model architectures, like LSTM and ConvLSTM \cite{conv_LSTM,pathak,helber2023crop}, Transformers \cite{vaswani2017attention,helber2024operational}, and Diffusion Models \cite{miranda_regdiff}. Moreover, to process multimodal input data, simple and advanced data fusion methods are used to handle different temporal, spatial, and spectral resolutions \cite{mena2024common,mena2024adaptive,miranda2024multi}. 
As ground truth data, regional \cite{srivastava2022winter} and field-level data \cite{cao2021wheat} are mainly used. Only a few studies use subfield-level yield data \cite{perich2023pixel}, mainly because of the high acquisition costs. Consequently, studies are limited to specific regions, crop types, and individual years \cite{pantazi2016wheat,you2017deep,gavahi2021deepyield,srivastava2022winter,wang2020winter}. 
Moreover, yield data is often affected by shifts in data distribution, driven by differences in management practices, environmental conditions, and climate variability \cite{ray2015climate}. Consequently, generalization to unknown years and regions often causes a severe performance reduction \cite{perich2023pixel,mena2024adaptive,miranda_drought} and, consequently, skepticism of deploying models into practice. \\ 
Only a few publicly available yield datasets exist, ready for training \gls{dl} models and to study yield prediction at large scale. 
A comparison of available datasets for yield prediction is provided in Tab. \ref{tab:YieldSAT_Comparision}. For instance, the \textit{SwissYield} \cite{perich2023pixel} dataset contains only 73 fields from a single crop type and country (Switzerland). In addition, the \textit{CropNet} \cite{lin2024open} dataset provides only regional-level yield data \cite{lin2024open,he2023physics}) for the US, with low temporal and spatial resolution. The data is coupled only with single bands and derivatives from \gls{s2}. \\
YieldSAT is, to our knowledge, the first public dataset that combines pixel-level yield maps with multimodal, globally-available \gls{eo} inputs across multiple crops and multiple countries ready for training \gls{dl} models. 
Moreover, only a few studies use \glspl{de} \cite{lakshminarayanan2017simple} for yield prediction \cite{shahhosseini2021corn}, primarily for uncertainty estimation. Exploring \glspl{de} to explore the impact of distribution shifts, such as evidenced in \cite{izmailov2021bayesian,wilson2020bayesian}, is an open challenge. 

\section{Dataset Overview}
\begin{table}[!tb]
    \centering
    \caption{\label{table:YieldSAT_Comparision}
        Comparison of the YieldSAT dataset with other crop yield prediction datasets. 
    }
    \resizebox{\columnwidth}{!}{
        \begin{tabular}{l|cccccllcc}
\hline
\textbf{Dataset}&
  \textbf{Countries} &  \textbf{Crops} &  \textbf{Years} &  \textbf{Fields}  & \textbf{Pixel-Level} & \multicolumn{2}{c}{\textbf{Resolution (Optical)}} & \textbf{Features} & \textbf{Curated}  \\ 
  \cmidrule(lr){7-8} 
                  &                  &         &   &  &    &   Spatial   & Temporal & \\ \hline

SwissYield \cite{perich2023pixel}  & 1         & 2          & 2017–2021           &   73                       & \cmark     &     $\SI{10}{m} $       & \textbf{$\sim$ 5 days}         & 14     &    \xmark                    \\
CropNet  \cite{lin2024open}        & 1         & \textbf{4}        & 2017-2022           &   0                       & \xmark            & $\SI{9}{km} $            & $\sim$ 14 days         & 13     &    \xmark                   \\
\textbf{YieldSAT (ours)}           & \textbf{4} & \textbf{4} & \textbf{2016-2024}  &  \textbf{2,173}   & \cmark     &  \textbf{$\SI{10}{m}$} & \textbf{$\sim$ 5 days} & \textbf{72} & \cmark \\ \hline
\end{tabular}
    } 
    \label{tab:YieldSAT_Comparision}
\end{table}
\begin{table}[!tb]
    \centering
    \caption{Overview of the available ground truth yield data. In total, 2,173 labeled fields (yield images) are available.}
    \resizebox{\linewidth}{!}{
        \begin{tabular}{l|cccccccccc}
\hline
 \textbf{\textit{YieldSAT}}  &     \multicolumn{5}{c}{\textbf{Fields}}  &  \textbf{Pixels} &    \multicolumn{2}{c}{\textbf{Area (ha)}}  & \textbf{Years} & \textbf{S2 Images} \\ 
    \cmidrule(lr){2-6} 
    \cmidrule(lr){8-9} 
     &  Corn &  Rapeseed &  Soybean & Wheat & Total &   &  Total  &  Average &  \\ \hline

Argentina & 185 &   \xmark  & 440 & 126 & 751 & $\sim$5.3 M & $\sim$55,898 &  74.3  & 2017-2024 & 42,931\\
Brazil    & 118 &   \xmark  & 293 & 140 & 551 & $\sim$4.2 M & $\sim$43,125 & 78.2 &  2017–2024 & 19,308 \\
Uruguay    & \xmark &   \xmark  & 572 & \xmark & 572 & $\sim$2.1 M & $\sim$32,804 &  57.3 &  2018–2022 & 24,318\\
Germany   &  \xmark   & 111 &  \xmark   & 188 & 299 & $\sim$0.6 M & $\sim$6,460 &   21.6  &   2016–2022 & 26,998\\ \hline
\textbf{Summary} &
  \multicolumn{1}{c}{\textbf{303}} &
  \multicolumn{1}{c}{\textbf{111}} &
  \multicolumn{1}{c}{\textbf{1,305}} &
  \multicolumn{1}{c}{\textbf{454}} &
  \multicolumn{1}{c}{\textbf{\underline{2,173}}} &  
  \multicolumn{1}{c}{\textbf{$\sim$12.2 M }} &
  \multicolumn{1}{c}{\textbf{$\sim$138,288}}&
  \multicolumn{1}{c}{\textbf{57.8}} &
  \multicolumn{1}{c}{\textbf{2016-2024}} & \textbf{113,555}\\ \hline
\end{tabular}
 }     
    \label{tab:yield_data}
\end{table}
\begin{figure}[!t]
    \centering
    \includegraphics[width=\columnwidth]{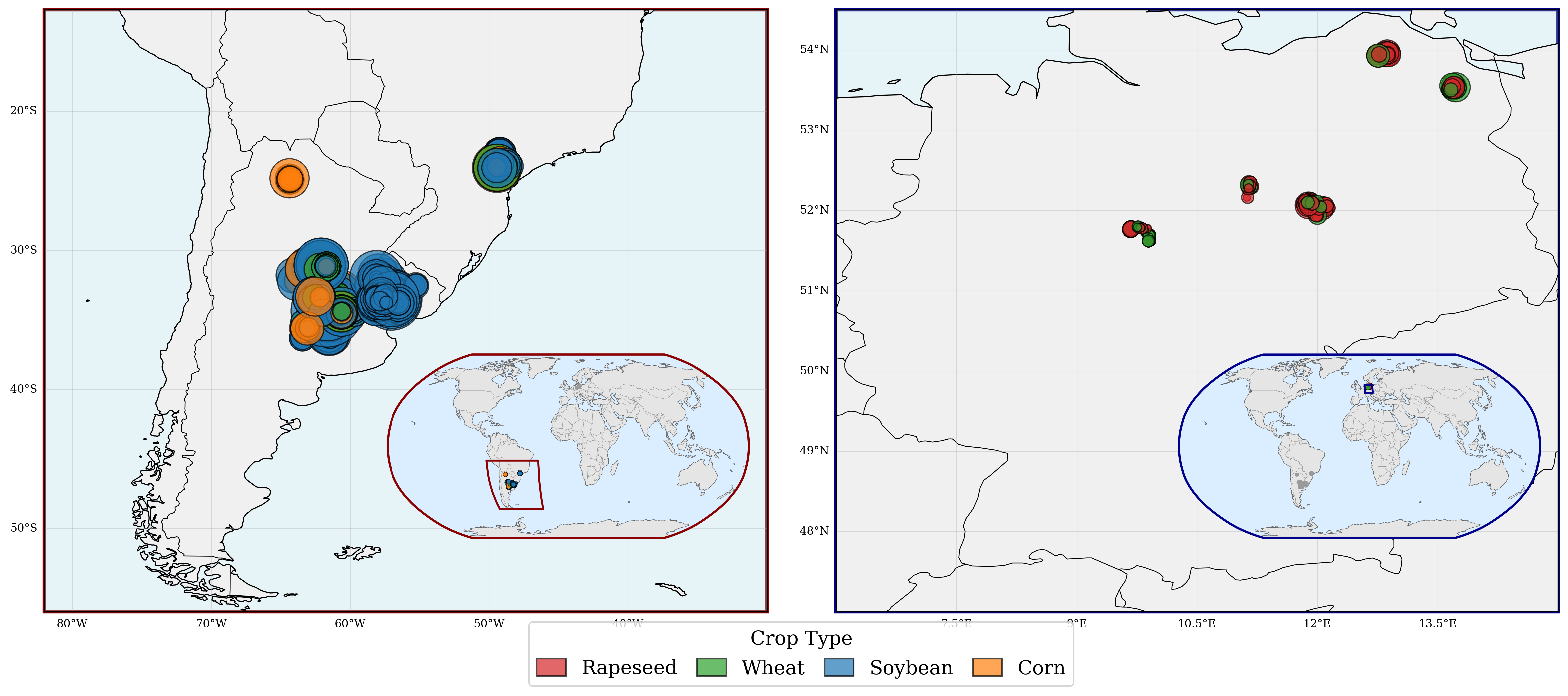}
    \caption{Spatial distribution of the collected yield data for each region and crop type. The yield data is colored by crop type. Left: South America, right: Europe. The marker size indicates the field size. (Map source: \cite{Cartopy}) 
    }
    \label{fig:yield_map_spatial}
\end{figure}


The \textit{YieldSAT} dataset is a multimodal dataset comprising high-resolution ($\SI{10}{m} \times \SI{10}{m}$) yield data, coupled with multimodal \gls{eo} data sources. The dataset spans a large area across Argentina, Brazil, Uruguay, and Germany. The available countries are major contributors to the global food production of the selected crop types \cite{owid_agricultural_production}. In total, the data covers a labeled area of approximately $\SI{138288}{ha}$ ($\sim\SI{1384}{km^2}$) with crop types of soybean (\textit{Glycine max L.}), corn (\textit{Zea mays}), rapeseed (\textit{Brassica napus subsp. napus}), and wheat (\textit{Triticum aestivum}). In Fig. \ref{fig:yield_map_spatial}, the geographic locations of the available fields are displayed, colored by crop type. 
Note that the availability of crop types varies by country. For example, Germany includes rapeseed and wheat, whereas Argentina provides data for corn, soybeans, and wheat. 
The dataset spans nine years, from 2016 to 2024, thereby covering large climate and yield variability. In total, 2,173 labeled fields are available. Soybean is the most represented crop, with 1,305 fields, while rapeseed is the least represented, with 111 fields. However, notable differences in the field size exist between countries. For example, Brazil exhibits the largest average field size, with $\SI{78.8}{ha}$. In contrast, Germany has smaller field sizes, with an average of $\SI{26.6}{ha}$. The average field size is $\SI{57.8}{ha}$ across the entire dataset. 
Altogether, the dataset includes approximately 12.2 million yield samples (labeled pixels) with $\SI{10}{m}$ resolution each.
A detailed description of the available ground truth data is depicted in Tab. \ref{tab:yield_data}. Note that pixel counts are nominal and yield measurements are inherently spatial autocorrelated due to environmental factors. \\
Each field is coupled with multimodal data sources, including multispectral satellite imagery from  \gls{s2}, weather data, soil data, and topography information. In total, 72 features are available for each sample. 
\section{Data Collection} \label{sec:dataset}
The data collection consists of (1) the ground truth yield data collection, (3) preprocessing, and (2) \gls{eo} data collection. 
\subsection{Yield Data Collection}
\textit{YieldSAT} contains subfield-level yield data from combine harvesters, collected at high spatial resolution. During harvest, a combine harvester equipped with yield monitors drives through the field and collects georeferenced data points as point vector data at a consistent frequency. 
Each data point contains various information, such as the geographic coordinate (latitude and longitude) of each measurement, the amount of wet yield, and the moisture content. 
Together, all data points form a yield map, a collection of data points in vector format for a single field. A schematic overview of the yield collection and a yield map is given in Fig. \ref{fig:Yieldmap_with_data}.
Yield maps provide valuable information about the spatial variability and productivity of a single field and enables farmers to identify productivity zones, estimate yield quality, and quantify damages and losses, and serves as a foundation for future activities.
\subsection{Yield Data Preprocessing}\label{sec:preprocessing}
Raw yield data is commonly collected as georeferenced point vector data in the \textit{shapefile} format. Still, many other data formats exist that can store geospatial yield data. Therefore, if the data is not provided in the shapefile format, a format conversion is performed as a first step. Moreover, yield maps are manually inspected and curated. The curation and other metadata are stored for each yield map, providing information on data quality, location, and other insights relevant to training yield prediction models. Only yield data was considered for further processing that appeared realistic to agricultural experts. 
Nevertheless, combine harvester yield data is heavily inhomogeneous across farmers, regions, and countries due to the use of different machines, languages, units, and management practices and therefore requires careful data preprocessing \cite{leroux2018general,sanchez2023influence}. 
To harmonize the data, a standardized preprocessing pipeline is used. This includes automatic and semi-automatic translation of feature naming over various languages and conversion to the metric system. 
Additionally, automatic transformation from the Geographic WGS84 to a projected UTM coordinate reference system is performed. \\
Combine harvester data is often mis-calibrated and associated with sensor errors, positioning inconsistencies, missing information, delays between the grain collection and measurement event, and focuses during turns \cite{talaviya2020implementation}. 
Consequently, removing erroneous values related to position, timestamp, yield, moisture, and inactive harvesters is essential to improve data quality and prevent misleading outcomes \cite{leroux2018general}. 
For this, based on expert rules, zero yield points and biologically infeasible points are removed. 
This includes crop-specific maximum yield values. In addition, data points are filtered by three standard deviations ($\pm 3\sigma$), following \cite{sanchez2023influence}.
Finally, the scaled yield (i.e., dry yield) is calculated based on the provided wet yield, adjusted to a fixed standard moisture, as  $y_s = y_w * (1 - m_m / 1 - m_s)$,
where $y_s$ is the scaled yield (dry yield), $y_w$ is the wet yield, $m_m$ is the measured moisture, and $m_s$ is the standard moisture. 
The Appendix \ref{tab:yield_threshold} gives maximum yield values and the standard moisture. The scaled yield is calculated because it is less affected by measurement noise (weather and time of harvest). Additionally, the scaled (dry) yield is the true indicator of the grain output used to estimate revenue potential for farmers, traders, and crop insurances.
\subsection{Earth Observation Data Collection}
We acquired \gls{eo} data for every yield map based on a stringent selection criteria: (1) demonstrated or theoretically established influence on crop development and yield, (2) open and freely accessible, (3) global coverage, and (4) high spatial resolution if possible. 
\subsubsection{Optical \& Multispectral Satellite Imagery}
\begin{table}[!tb]
    \centering
    \caption{Overview of all available data modalities in the \textit{YieldSat} dataset and their characteristics.}
    \resizebox{.9\columnwidth}{!}{
\begin{tabular}{cclcc}
\hline
\multicolumn{1}{l}{\textbf{Modality}} &
  \multicolumn{1}{l}{\textbf{Source}} &
  \textbf{Product} &
  \multicolumn{2}{c}{\textbf{Resolution}} \\ 
  \cmidrule(lr){4-5} 
                  &         &   &  Spatial   & Temporal \\ \hline

\multirow{12}{*}{Multispectral} &
  \multirow{12}{*}{Sentinel-2 L2A} &
  B01 - Coastal Aerosol &
  \multicolumn{1}{c}{$\SI{60}{m} $} &
  \multirow{12}{*}{$\thicksim$ 5 days} \\
  
 &  & B02 - Blue                              & \multicolumn{1}{c}{$\SI{10}{m}$} &  \\
 &  & B03 - Green                             & \multicolumn{1}{c}{$\SI{10}{m} $} &  \\
 &  & B04 - Red                               & \multicolumn{1}{c}{$\SI{10}{m}$} &  \\
 &  & B05 - Red Edge 1                        & \multicolumn{1}{c}{$\SI{20}{m} $} &  \\
 &  & B06 - Red Edge 2                        & \multicolumn{1}{c}{$\SI{20}{m} $} &  \\
 &  & B07 - Red Edge 3                        & \multicolumn{1}{c}{$\SI{20}{m} $} &  \\
 &  & B08 - NIR                               & \multicolumn{1}{c}{$\SI{10}{m} $} &  \\
 &  & B8A - Narrow NIR                        & \multicolumn{1}{c}{$\SI{20}{m} $} &  \\
 &  & B09 - Water vapour                      & \multicolumn{1}{c}{$\SI{60}{m} $} &  \\
 &  & B11 - SWIR 1                            & \multicolumn{1}{c}{$\SI{20}{m} $} &  \\
 &  & B12 - SWIR 2                            & \multicolumn{1}{c}{$\SI{20}{m} $} &  \\ 
 &  & Scene Classification Layer       & \multicolumn{1}{c}{$\SI{20}{m} $} & \\ \hline
\multirow{4}{*}{Weather} &
  \multirow{4}{*}{Era5 \cite{hersbach2020era5}} &
  Max Temperature &
  \multirow{4}{*}{$\SI{30}{km}$} &
  \multirow{4}{*}{Daily} \\
 &  & Mean Temperature                        &                          &  \\
 &  & Min Temperature                         &                          &  \\
 &  & Total Precipitation                     &                          &  \\ \hline
\multirow{8}{*}{Soil} &
  \multirow{8}{*}{SoilGrids \cite{poggio2021soilgrids}} &
  Soil Organic Carbon &
  \multirow{8}{*}{$\SI{250}{m}$} &
  \multirow{13}{*}{Static} \\
 &  & Nitrogen                                &                          &  \\
 &  & Cation Exchange Capacity                &                          &  \\
 &  & Clay                                    &                          &  \\
 &  & Silt                                    &                          &  \\
 &  & Sand                                    &                          &  \\
 &  & pH                                      &                          &  \\
 &  & Coarse fragments &                          &  \\ \cline{1-4}
\multirow{5}{*}{Topography} &
  \multirow{5}{*}{SRTM \cite{farr2000shuttle}} &
  Topography (DEM) &
  \multirow{5}{*}{$\SI{30}{m}$} &
   \\
 &  & Slope                                   &                          &  \\
 &  & Curvature                               &                          &  \\
 &  & TWI                                       &                          &  \\
 &  & Aspect                                  &                          &  \\ \hline
\end{tabular}
    } 
    \label{tab:data_modalities}
\end{table}
For each yield map, \gls{s2} Level-2A multispectral satellite imagery, including all 13 spectral bands, was acquired for the period between seeding and harvesting. The yield map boundary serves as a geo-reference for the image acquisition process, resulting in a time series with a temporal resolution of approximately one image every five days. Low-resolution bands were nearest-neighbor upsampled to $\SI{10}{m}$ to ensure a uniform spatial resolution across all bands. 
The \gls{s2} bands are preserved in the original form, and no specific indices are calculated (e.g., NDVI, NDWI). This increases the flexibility of the dataset for downstream tasks. 
Additionally, the \textit{Scene Classification Layer (SCL)} was acquired for each time step, providing per-pixel class information for the \gls{s2} product at $\SI{20}{m}$ resolution. In total, the SCL provides 12 class labels, including “vegetated,” “non-vegetated,” and “clouded”. 
\subsubsection{Additional Data Modalities}
We further provide \glspl{adm} (weather, soil, and topography) to complement and compensate for inconsistencies and shortcomings of the \gls{s2} data. For example, \gls{s2} often suffers from missing time steps (e.g., due to cloud occlusion), which introduce uncertainty into the model \cite{miranda2025analysis}. 
Weather data for each field were derived from the ECMWF Reanalysis (ERA5) program \cite{hersbach2020era5} between seeding and harvesting, at a daily resolution. Soil data was acquired from the \textit{SoilGrids} archive in $\SI{250}{m}$ resolution \cite{poggio2021soilgrids}. Topography data was acquired from the SRTM mission \cite{farr2000shuttle} in $\SI{30}{m}$ resolution. For soil and topography data, raster images are generated for each feature and upsampled to $\SI{10}{m}$ resolution using cubic spline interpolation to match the \gls{s2} image resolution. For soil, all soil properties are sampled at depths of 0-5, 5-15, 15-30, 30-60, 60-100, and 100-200 cm with the uncertainty provided for every layer. 
For the topography data, the \textit{RichDEM} \cite{RichDEM} library was used for feature engineering to derive additional features, including aspect, curvature, slope, and the \textit{Topographic Wetness Index (TWI)}. The TWI is derived following \cite{kopecky2021topographic}. 
A detailed overview of the available data modalities is given in Tab. \ref{tab:data_modalities}.
\subsection{Rasterization \& Data Quality}
\begin{figure*}[!tb]
    \centering
    \includegraphics[width=.95\textwidth]{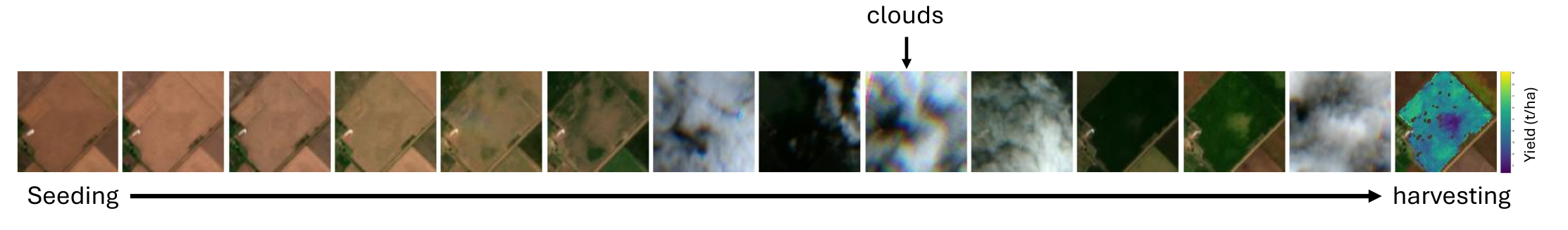}
    \captionof{figure}{ Example satellite time series of a soybean field from seeding (left) to harvesting (right). The last image shows the collected yield at harvest.}
    \label{fig:s2_time_series}
\end{figure*}
\begin{figure}[!tb]
    \centering
    \includegraphics[width=\columnwidth]{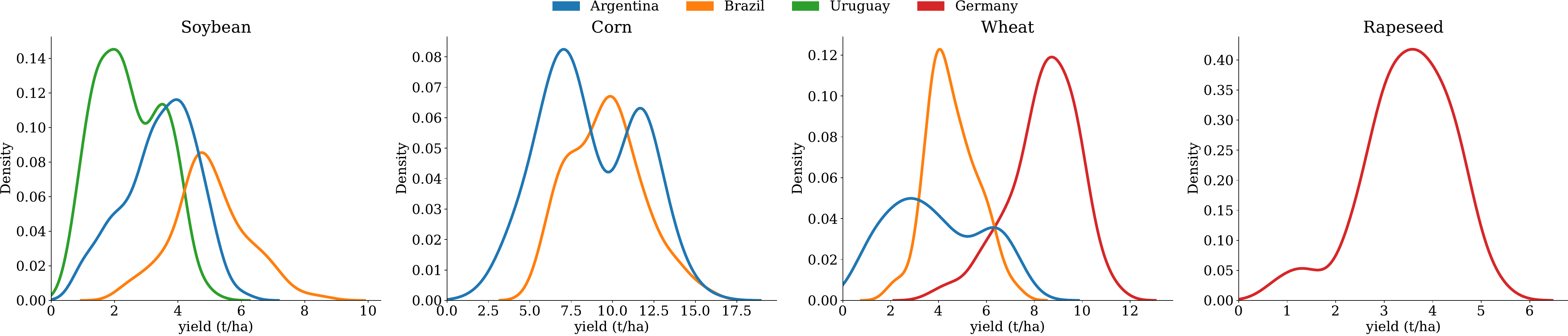}
    \caption{Yield data distribution plots for each crop type and country, averaged per field.
}
    \label{fig:YieldSat_distribution}
\end{figure}
\begin{figure}[!tb]
    \centering
    \includegraphics[width=.99\linewidth]{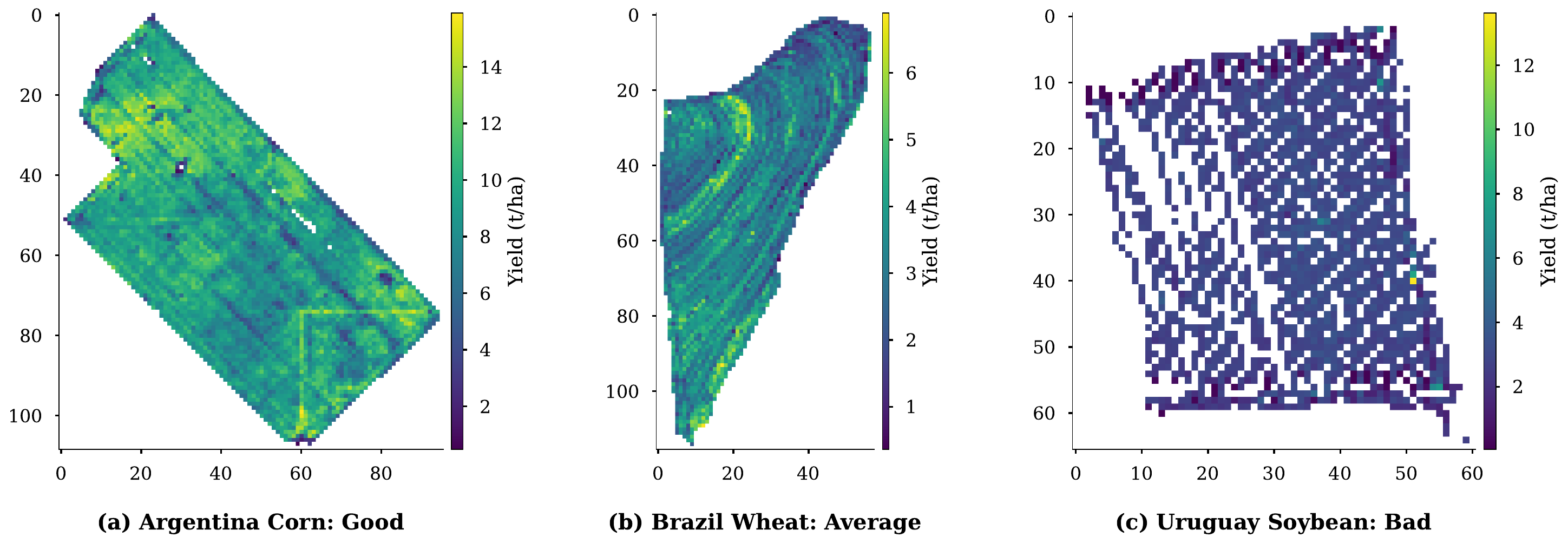}
    \caption{Examples of rasterized yield maps with the curation for each quality level (\textit{good}, \textit{average}, and \textit{bad}).}
    \label{fig:yield_map_quality}
\end{figure}
\begin{figure}[!tb]
    \centering
    \includegraphics[width=.99\linewidth]{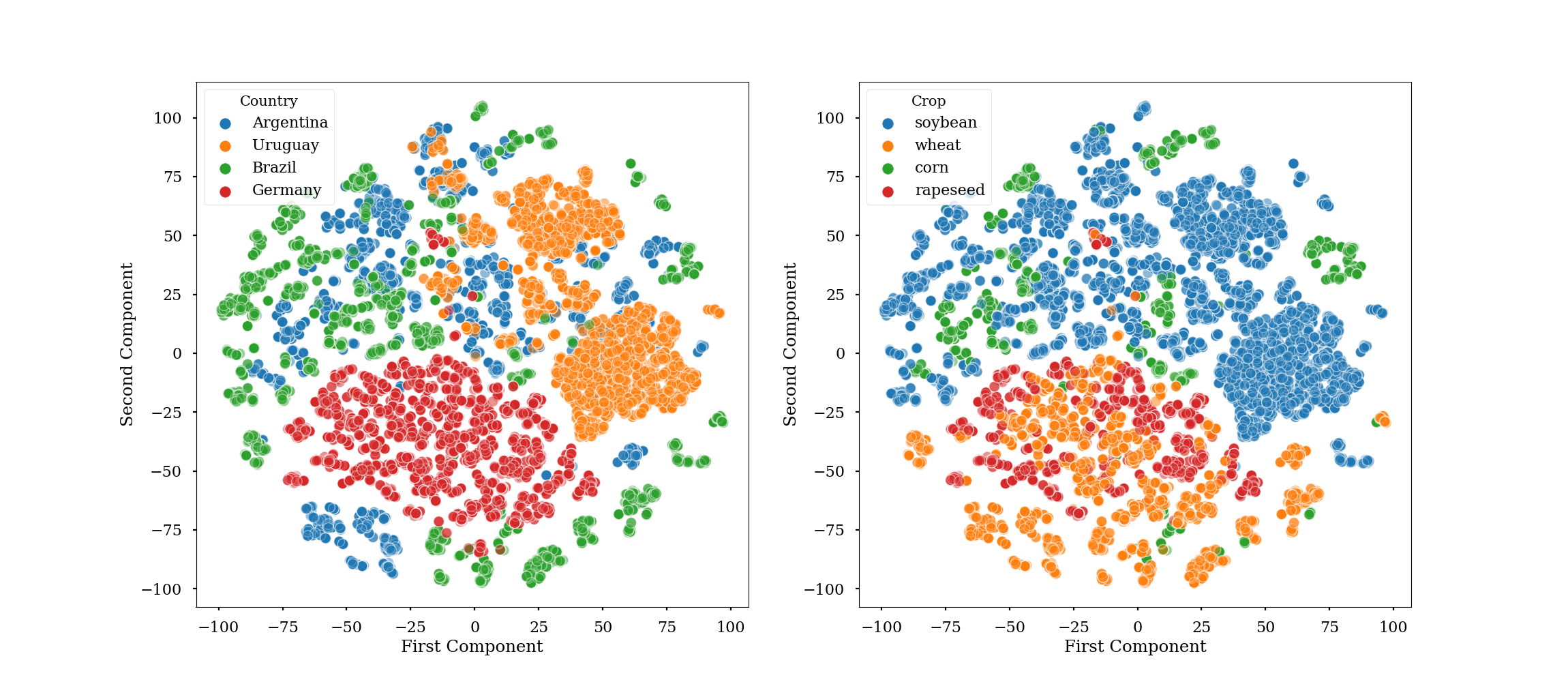}
    \caption{t-SNE plot of the \gls{s2} surface reflectance colored by countries (left) and crops (right).}
    \label{fig:TSNE_S2_All_combined}
\end{figure}
To align the \gls{s2} imagery and the \glspl{adm} with the yield data, yield maps are rasterized so that each \gls{s2} product pixel aligns spatially with the corresponding yield data pixel across the entire time series. For this, a rasterization grid derived from \gls{s2} imagery at $\SI{10}{m}$ resolution is overlaid on the yield map. 
All yield points within a given pixel are averaged, yielding a rasterized yield image with the same spatial resolution as the \gls{s2} data. Consequently, each pixel in the \gls{s2} product is spatially aligned with a pixel in the target yield image. An example time series of \gls{s2} imagery with the final rasterized yield image is depicted in Fig. \ref{fig:s2_time_series}. 
Averaging per pixel follows state-of-the-art processing for harvester data and preserves both field-level means and subfield patterns. Pixels without vector points are masked and not used for training. 
Nevertheless, rasterization can introduce variable support sizes for each pixel due to harvester path density, swath width, speed, logging frequency, and positional delay, which are often correlated in space. This can impact yield modeling and introduce spatially correlated uncertainty. To account for variable support sizes and spatially correlated uncertainty in future research, we additionally provide label information for each image: (i) the number of yield points and (ii) the standard deviation, both per $\SI{10}{m}$ pixel. \\ 
Every yield map contains a quality label from manual expert labeling.  Fig. \ref{fig:yield_map_quality} presents randomly selected rasterized yield maps for each quality level (\textit{good}, \textit{average}, and \textit{bad}), highlighting the varying quality levels within the dataset. Low-quality yield maps are often characterized by sparse data distribution, spatial misalignment, or erroneous measurements, such as unrealistically high yield values. Additionally, artifacts may be present, including patterns caused by harvester turns or delays in the measurement process. A more in-depth analysis of the rasterization, data quality, and support sizes is provided in the Appendix (see \ref{sec:quality_support}). \\
In summary, the \textit{YieldSAT} dataset is highly diverse, with distinct yield distributions for each crop type and country. For instance, corn and wheat tend to exhibit the highest yields, with a broad spread, while soybeans and rapeseed generally show lower yields.  
This is depicted in Fig. \ref{fig:YieldSat_distribution}. More importantly, the yield data distributions are significantly different between countries and crops and between years and regions for a single crop type and country (p-values $<0.0001$, see Appendix \ref{sec:distribution_shifts_country_crop}). 
Additionally, we observe high diversity in the patterns of the data modalities (e.g., surface reflectance for the \gls{s2} time series) between countries and crop types. 
In Fig. \ref{fig:TSNE_S2_All_combined} a t-Distributed Stochastic Neighbor Embedding (t-SNE) \cite{maaten2008visualizing} of the surface reflectance of the \gls{s2} time series is shown, colored by countries (left) and crops (right). The plot shows that the surface reflectance differs significantly between countries and crops, increasing the difficulty of generalizing between different environments. 

\subsection{Data Availability}
The final dataset is available in two formats: (1) a preprocessed version using an input fusion strategy, described in \cite{helber2023crop,pathak}. Here, the input modalities are aligned via concatenation and temporal and spatial repetition, resulting in a unified time series of 24 time steps, encompassing all available data modalities.  
The dataset is stored in the \textit{xarray} data format \cite{hoyer2017xarray} and ready to train \gls{dl} models. (2) A version where each field is stored jointly with the described input modalities and further meta information. This version enables the development of advanced \gls{dl} models by offering high flexibility. For both data formats, the preprocessing and cleaning were done as described. 
Further information about the data, limitations, legal and ethical considerations, and data distribution is given in the data sheet in the Appendix (see \ref{sec:datasheet}).

\section{Experiments}

\begin{table*}[!tb]
    \centering
    \caption{Results for the RMSE (t/ha) ($\downarrow$) and the $R^2$-score ($\uparrow$) for different models and datasets. The best model is highlighted in bold, and the best overall score is underlined. ARG = Argentina, BRA = Brazil, GER = Germany, URG = Uruguay. C = corn, R = rapeseed, S = soybean, W = wheat. }
    \resizebox{.99\linewidth}{!}{
\begin{tabular}{cclllllllllll|llllllllllll}
\hline
\multicolumn{3}{c}{\multirow{2}{*}{\textbf{Evaluation}}} &
  \multicolumn{10}{c|}{\textbf{Field-Level}} &
  \multicolumn{12}{c}{\textbf{Subfield (Pixel)-Level}} \\ \cline{4-25} 
\multicolumn{3}{c}{} &
  \multicolumn{2}{c}{\textbf{ARG-S}} &
  \multicolumn{2}{c}{\textbf{BRA-C}} &
  \multicolumn{2}{c}{\textbf{GER-R}} &
  \multicolumn{2}{c}{\textbf{GER-W}} &
  \multicolumn{2}{c|}{\textbf{URG-S}} &
  \multicolumn{2}{c}{\textbf{ARG-S}} &
  \multicolumn{2}{c}{\textbf{BRA-C}} &
  \multicolumn{2}{c}{\textbf{BRA-S}} &
  \multicolumn{2}{c}{\textbf{GER-R}} &
  \multicolumn{2}{c}{\textbf{GER-W}} &
  \multicolumn{2}{c}{\textbf{URG-S}} \\ \hline
\textbf{Modalities} &
  \multicolumn{1}{c}{\textbf{Fusion Method}} &
  \multicolumn{1}{c}{\textbf{Model}} &
   $R^2$ &
  RMSE &
   $R^2$ &
  RMSE &
   $R^2$ &
  RMSE &
   $R^2$ &
  RMSE &
   $R^2$ &
  RMSE &
   $R^2$ &
  RMSE &
   $R^2$ &
  RMSE &
   $R^2$ &
  RMSE &
   $R^2$ &
  RMSE &
   $R^2$ &
  RMSE &
   $R^2$ &
  RMSE \\ \hline
\multirow{4}{*}{S2} &
  \xmark &
  3D-ConvLSTM \cite{helber2024operational} &
  0.79   &
  0.55   &
  0.82   &
  0.74   &
  0.81  &
  0.58   &
  0.65   &
  1.12   &
  0.77   &
  0.51  &
  0.65   &
  0.90   &
  0.45   &
  2.13   &
  0.39   &
  0.94   &
  0.49  &
  1.20   &
  0.34   &
  2.40   &
  0.41   &
  1.22   \\
 &
  \xmark &
  3D-LSTM \cite{miranda2024multi} &
  0.77   &
  0.58   &
  0.82   &
  0.74   &
  0.82   &
  0.57   &
  0.54   &
  1.28   &
  0.73   &
  0.56   &
  0.65   &
  0.90   &
  0.46   &
  2.13   &
  0.39   &
  0.93   &
  0.48   &
  1.20   &
  0.30  &
  2.46   &
  0.39   &
  1.23   \\
 &
  \xmark &
  LSTM \cite{pathak} &
  0.72   &
  0.64   &
  0.75  &
  0.88   &
  0.62   &
  0.83   &
  0.55   &
  2.60   &
  0.66   &
  0.62   &
  0.60   &
  0.96   &
  0.42   &
  2.20   &
  0.34   &
  0.98   &
  0.36   &
  1.33   &
  0.32   &
  2.47   &
  0.37   &
  1.26   \\
 &
  \xmark &
  Transformer \cite{helber2023crop} &
  0.73   &
  0.63   &
  0.79   &
  0.82   &
  0.75   &
  0.67   &
  0.56   &
  1.26   &
  0.72  &
  0.56  &
  0.62   &
  0.94  &
  0.44   &
  2.16   &
  0.38   &
  0.94   &
  0.44  &
  1.25   &
  0.32   &
  2.43   &
  0.38   &
  1.24   \\ \hline
\multirow{6}{*}{S2 + ADM} &
  \multirow{4}{*}{Input Fusion} &
  3D-ConvLSTM \cite{helber2024operational} &
  0.82   &
  0.52   &
  0.83   &
  0.72   &
  0.78   &
  0.63   &
  0.70   &
  1.03   &
  0.78   &
  0.50   &
  0.68   &
  0.87   &
  0.46   &
  2.12   &
  0.38   &
  0.94   &
  0.42   &
  1.27   &
  0.39   &
  2.30   &
  0.41   &
  1.22   \\
 &
   &
  3D-LSTM \cite{miranda2024multi} &
  0.76  &
  0.59  &
  0.84 &
  0.71  &
  0.81  &
  0.59  &
  0.62  &
  1.17  &
  0.76  &
  0.54  &
  0.64  &
  0.91  &
  0.46  &
  2.12  &
  0.41  &
  0.93  &
  0.49 &
  1.20  &
  0.37  &
  2.34  &
  0.40  &
  1.22  \\
 &
   &
  LSTM \cite{pathak} &
  0.72   &
  0.64   &
  0.81  &
  0.78   &
  0.81   &
  0.58   &
  0.63   &
  1.16   &
  0.72   &
  0.56   &
  0.59   &
  0.98   &
  0.43   &
  2.18   &
  0.33   &
  0.99   &
  0.47   &
  1.21   &
  0.35   &
  2.38  &
  0.39   &
  1.23   \\
 &
   &
  Transformer \cite{helber2023crop}&
  0.72   &
  0.64   &
  0.79   &
  0.81   &
  0.76   &
  0.65   &
  0.61   &
  1.19   &
  0.73   &
  0.55   &
  0.58   &
  0.98   &
  0.44   &
  2.17   &
  0.37   &
  0.96   &
  0.45   &
  1.24   &
  0.38   &
  2.32   &
  0.39   &
  1.23   \\ \cline{2-25} 
 &
  \multirow{2}{*}{Feature Fusion}&
  AFF  \cite{miranda2024multi}  &
  \underline{\textbf{0.84}} &
  0.49  &
  \underline{\textbf{0.84}} &
  0.70   &
  0.80   &
  0.60   &
  0.74   &
  0.96   &
  0.81   &
  \underline{\textbf{0.46}}  &
  \underline{\textbf{0.73}}  &
  \underline{\textbf{0.79}} &
  0.46   &
  2.12   &
  0.44   &
  0.90   &
  0.49   &
  1.20   &
  0.44   &
  2.20   &
  0.43   &
  1.19   \\
 &
   &
  MMGF \cite{mena2024adaptive} &
  0.82   &
  0.51   &
  0.76   &
  0.86   &
  0.75   &
  0.68   &
  0.77   &
  0.90   &
  0.75   &
  0.53   &
  0.70   &
  0.84   &
  0.42   &
  2.19   &
  0.42   &
  0.92   &
  0.44   &
  1.26   &
  0.44   &
  2.21   &
  0.40   &
  1.22   \\ \hline
\end{tabular}

    } 
    \label{tab:benchmark}
\end{table*}

\begin{figure}[!tb]
    \centering
    \includegraphics[width=.9\columnwidth]{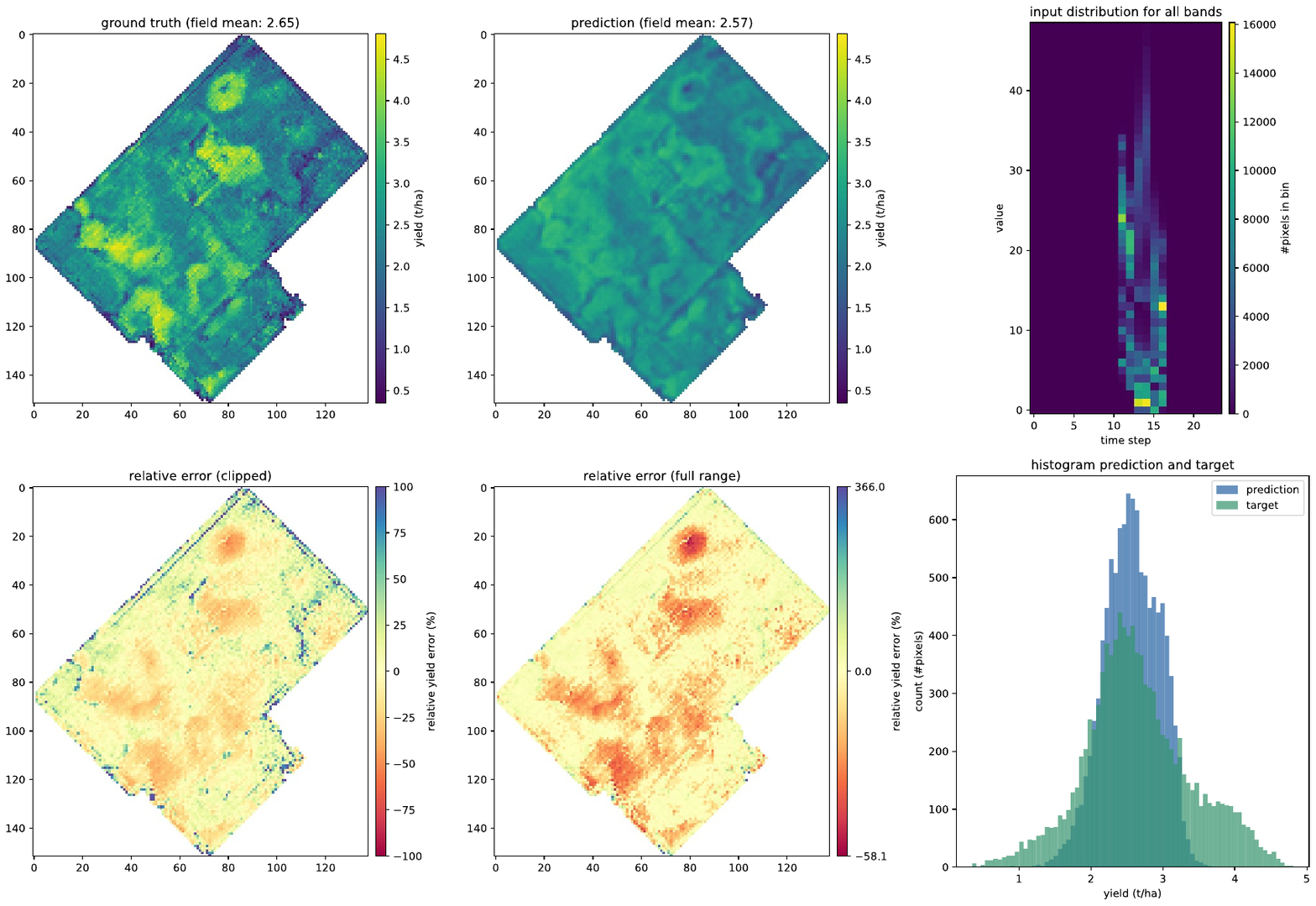}
    \caption{Qualitative results for a single soybean field from Argentina, generated with the 3D-LSTM model. Top left to bottom right: ground truth yield map, predicted yield map, prediction over target plot, input distribution over time, clipped relative pixel-wise error (100\%), relative pixel-wise error (full range), histogram of predicted (blue) and target (green) values. 
    }
    \label{fig:prediction_example}
\end{figure}
In this section, we demonstrate the potential of large-scale, high-resolution crop yield prediction with different \gls{dl} models using previously published methods. 
All models are trained at the pixel level, as in the state-of-the-art approaches, treating each pixel as an independent sample. 
Since pixels are highly autocorrelated due to e.g., shared soil properties, topography, management practices, or microclimate, some methods include spatial neighborhood information by processing a pixel with its surroundings using a 3D-CNN block or a ConvLSTM \cite{shi2015convolutional} approach. Such approaches account for spatial dynamics and spatial correlations within the field (image) \cite{miranda2024multi}. 
To prevent overfitting and information leakage, models were trained using a stratified, grouped 10-fold cross-validation (cv), where pixels are grouped by field and stratified by region. This is done to ensure that pixels from the same field are either in the training or testing set.
The metrics are presented as the average across the folds and are used for relative model ranking. 
All models are evaluated based on their ability to predict yield at both the subfield (i.e., pixel) and field levels. For field-level performance, all pixels in the same field are averaged and compared with the field's averaged ground truth. \\
Results are reported on subsets per country and crop due to the heterogeneity of the dataset, which is how the data will be used in practice. \\
We compare models trained only on \gls{s2} data and models that further incorporate \glspl{adm} using a simple input fusion method \cite{pathak} and advanced fusion techniques \cite{mena2024common}. 
\subsection{Benchmark Results}
An overview of benchmark results for different model architectures and selected subsets is provided in Tab. \ref{tab:benchmark} for the $R^2$-score and for the RMSE in tons/hectare (t/ha). The presented subsets contain key characteristics (crops/countries/quality) while providing key insights and maintaining readability. A full replication of all available subsets is provided in the Appendix (see \ref{sec:further_results}). \\
Notably, the results depend strongly on the country and crop type, likely due to differences in the quality of the ground truth data. For instance, soybean in Argentina (ARG-S) exhibits high performance across all models, with a maximum $ R^2$ score of 0.84 and an RMSE of 0.49 t/ha. Additionally, there are significant differences between subfield- and field-level performance, with the field-level exhibiting consistently higher scores, especially for advanced fusion methods. 
In general, modeling spatial correlation using 3D-CNN blocks (3D-LSTM, 3D-ConvLSTM, AFF) significantly improves the performance compared to modeling each pixel independently. Such approaches even perform considerably well when using only \gls{s2} data as input. However, we emphasize that integrating \glspl{adm} commonly improves performance compared to using \gls{s2} data alone. Nevertheless, the benefit of \glspl{adm} depends on the model architecture and the fusion strategy. Related studies demonstrate that optical data, e.g., \gls{s2} is crucial for pixel-wise yield prediction \cite{miranda2024multi,mena2024adaptive,najjar2025explainability}. For instance, coupling spatial information (3D-LSTM, 3D-ConvLSTM) with an input fusion strategy even results in a performance reduction, which is related to the difference in temporal and spatial resolution of the input data \cite{miranda2024multi}. 
Consequently, coupling multimodal data with a more complex architecture requires advanced fusion methods to harmonize different spatial, temporal, and spectral resolutions. A qualitative example of a predicted field is shown in Fig. \ref{fig:prediction_example}, which depicts high subfield variability and a good match between the input and target distributions, while also highlighting areas of high pixel-wise error. 
\subsection{Distribution Shift \& Deep Ensembles}
\begin{table}[!tb]
    \centering
    \caption{
    Overview for crop yield prediction at the field level using temporal (Leave-One-Year-Out) and spatial splitting (Leave-One-Region-Out). All models, except the baseline, are defined as a \textit{Deep Ensemble} \cite{lakshminarayanan2017simple} with 5 ensemble members. All models were trained on \gls{s2} data only on subsets from Argentina. DE = Deep Ensemble. The best score is highlighted in bold.
    }
    \resizebox{.9\columnwidth}{!}{
\begin{tabular}{cl|ll|ll}
\hline
\multicolumn{2}{c|}{Evaluation}     & \multicolumn{2}{c|}{Leave-One-Year-Out} & \multicolumn{2}{l}{Leave-One-Region-Out} \\ \hline
\multicolumn{1}{l}{Dataset} & Model &  \multicolumn{1}{c}{\begin{tabular}[c]{@{}c@{}}RMSE ($\downarrow$)\\ t/ha\end{tabular}}                & \multicolumn{1}{c|}{\begin{tabular}[c]{@{}c@{}}$R^2$ ($\uparrow$)\\ -\end{tabular}}                &  \multicolumn{1}{c}{\begin{tabular}[c]{@{}c@{}}RMSE ($\downarrow$)\\ t/ha\end{tabular}}                 & \multicolumn{1}{c}{\begin{tabular}[c]{@{}c@{}}$R^2$ ($\uparrow$)\\ -\end{tabular}}                \\ \hline
\multirow{3}{*}{Soybean} & DE-LSTM       &   0.70 {\tiny $\pm$ \textcolor{black!50}{0.25}}  &    0.55 {\tiny $\pm$  \textcolor{black!50}{0.64}} & 0.65 {\tiny $\pm$ \textcolor{black!50}{0.20}} & 0.65 {\tiny $\pm$ \textcolor{black!50}{0.87}}    \\
                       & DE-3D-LSTM & \textbf{0.23}  {\tiny $\pm$ \textcolor{black!50}{0.11}} &  \textbf{0.63}  {\tiny $\pm$ \textcolor{black!50}{0.40}} & \textbf{0.19} {\tiny $\pm$ \textcolor{black!50}{0.14}}  &  \textbf{0.73} {\tiny $\pm$ \textcolor{black!50}{0.53}}  \\ 
                       & Baseline LSTM &   0.85  {\tiny $\pm$ \textcolor{black!50}{0.23}} &  0.50 {\tiny $\pm$ \textcolor{black!50}{0.50}} &  0.72 {\tiny $\pm$ \textcolor{black!50}{0.15}} &  0.64 {\tiny $\pm$ \textcolor{black!50}{0.15}}  \\ \hline
\multirow{3}{*}{Corn} & DE-LSTM       &  2.06  {\tiny $\pm$ \textcolor{black!50}{0.82}}  &  0.46 {\tiny $\pm$  \textcolor{black!50}{0.52}} & 1.81  {\tiny $\pm$ \textcolor{black!50}{0.86}} & 0.59 {\tiny $\pm$ \textcolor{black!50}{5.28}}    \\
                       & DE-3D-LSTM & \textbf{1.70}  {\tiny $\pm$ \textcolor{black!50}{0.61}} &  \textbf{0.63}  {\tiny $\pm$ \textcolor{black!50}{0.36}} & \textbf{1.36} {\tiny $\pm$ \textcolor{black!50}{0.83}}  &  \textbf{0.76} {\tiny $\pm$ \textcolor{black!50}{2.31}}  \\ 
                       & Baseline LSTM &   2.06  {\tiny $\pm$ \textcolor{black!50}{1.07}} &  0.46 {\tiny $\pm$ \textcolor{black!50}{0.52}} &  2.05
 {\tiny $\pm$ \textcolor{black!50}{0.75}} &  0.47 {\tiny $\pm$ \textcolor{black!50}{1.36}}  \\ \hline
\multirow{3}{*}{Wheat} & DE-LSTM       &  0.98   {\tiny $\pm$ \textcolor{black!50}{0.45}}  &  0.79   {\tiny $\pm$  \textcolor{black!50}{0.95}} & 1.11 {\tiny $\pm$ \textcolor{black!50}{0.47}} & 0.73 {\tiny $\pm$ \textcolor{black!50}{1.69}}    \\
                       & DE-3D-LSTM & \textbf{0.99}  {\tiny $\pm$ \textcolor{black!50}{0.35}} &  \textbf{0.79}  {\tiny $\pm$ \textcolor{black!50}{1.24}} & \textbf{0.98} {\tiny $\pm$ \textcolor{black!50}{0.36}}  &  \textbf{0.79} {\tiny $\pm$ \textcolor{black!50}{1.24}}  \\ 
                       & Baseline LSTM &   1.02  {\tiny $\pm$ \textcolor{black!50}{0.37}} &  0.77 {\tiny $\pm$ \textcolor{black!50}{1.49}} &  1.18 {\tiny $\pm$ \textcolor{black!50}{0.26}} &  0.70 {\tiny $\pm$ \textcolor{black!50}{0.38}}  \\\hline 
\end{tabular}

    } 
    \label{tab:loro_lofo}
\end{table}
\begin{figure}[!tb]
  \centering
  \includegraphics[width=\linewidth]{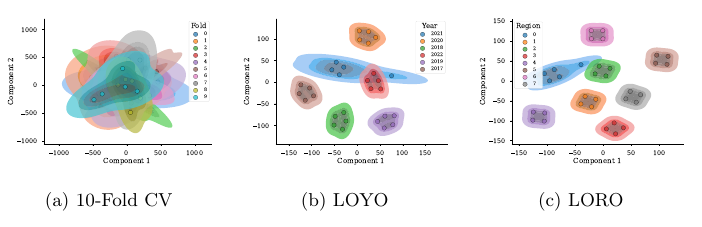}
  \caption{t-SNE of the network parameters for the Deep Ensemble model for different CV scenarios. Left: standard 10-CV, center: LOYO, right:  LORO. The weights are colored by fold.}
\label{fig:TSNE_Parameters}
\end{figure}
\begin{figure}[!tb]
  \centering
    \begin{subfigure}{.48\linewidth}
      \includegraphics[width=\linewidth]{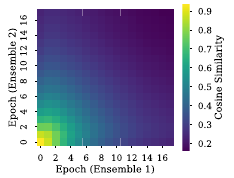}
    \end{subfigure}
    \begin{subfigure}{.48\linewidth}
      \includegraphics[width=\linewidth]{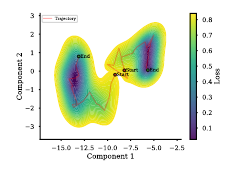}
    \end{subfigure}
  \caption{Visualization of the weight space diversity during model training. Left: Cosine similarity between two ensemble members during training. Right: PCA plot of the weight space during training, together with the loss. The trajectory in the weight space is highlighted in red from start to end. }
\label{fig:parameter_space_diversity}
\end{figure}
\begin{figure}[!tb]
  \centering
  \includegraphics[width=.99\columnwidth]{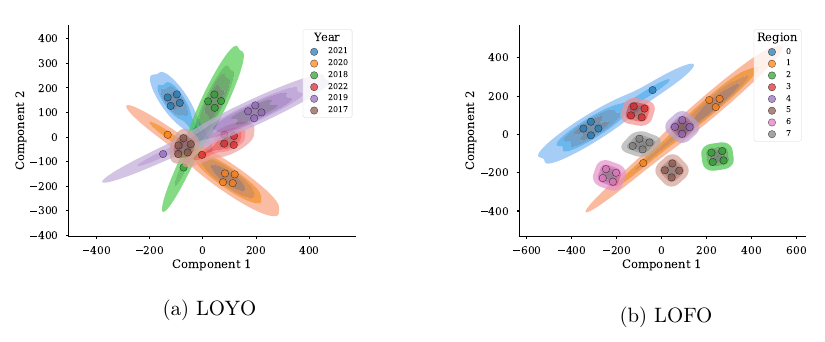}
  \caption{t-SNE of the network parameters for the Deep Ensemble model trained with prior knowledge (3D-LSTM) under distribution shift. Left: LOYO, right: LORO.}
\label{fig:TSNE_Parameters_iML}
\end{figure}
\gls{dl} models are susceptible to distribution shifts and exhibit overconfidence and degraded performance when exposed to such data \cite{NEURIPS2021_07ac7cd1,Ekim_2025_CVPR}. We already reported that the \textit{YieldSAT} dataset is affected by distribution shifts, even between years and regions for a single country (see Appendix \ref{sec:distribution_shifts_year_region}).
In this section, we explore the impact of distribution shift in \gls{dl} models using a real-world scenario, namely a \textit{Leave-One-Year-Out (LOYO)} and \textit{Leave-One-Region-Out (LORO)} CV experiments. A region is defined by a set of fields belonging to a single farmer or to a local data provider.
Subsequently, a model is evaluated on a held-out year or region to assess its generalization to unknown distributions. 
We focus on LOYO and LORO experiments, as this reflects how the dataset will be used in practice. Nevertheless, cross-crop and cross-country evaluations should be investigated in future work. In preliminary experiments, however, these settings led to model collapse without domain adaptation due to severe data heterogeneity. \\
To explore the impact of distribution shifts, we employ a \gls{de} approach \cite{lakshminarayanan2017simple}. 
Specifically, we evaluate an LSTM model in an ensemble setting and the 3D-LSTM \cite{miranda2024multi} that incorporates spatial correlations. This architecture was chosen because of its good performance in the earlier experiments and for its lightweight design compared to the advanced fusion methods, since training \glspl{de} is computationally expensive. Moreover, the model is trained solely on \gls{s2} data based on insights from our earlier findings. 
The results are shown on crops from Argentina only as the largest and best-curated subset (see Appendix \ref{sec:quality_support}) with strong year/region shifts, which provides statistical power for computationally expensive experiments and weight space analysis under real-world conditions. 
The results are presented in Tab. \ref{tab:loro_lofo} and confirm the hypothesis that model performance significantly decreases under distributional shift. 
For instance, for the LOYO experiment, an overall reduction in the $R^2$-score of 22 p.p. is reported compared to the standard CV experiments for soybean (see results for LSTM in Tab. \ref{tab:benchmark}). Likewise, for the LORO experiment, a severe reduction of 8 p.p. is shown. In the Appendix we show that the impact of distribution shifts is consistent over all model architectures and datasets (see Appendix \ref{sec:further_results}).
In contrast, the \gls{de} demonstrates improved performance compared to the baseline model in both settings, as shown in Tab. \ref{tab:loro_lofo}.  
In the LOYO setting, the \gls{de} improves by 5 p.p. $R^2$ over the baseline model for soybean. 
Likewise, the \gls{de} model exhibits a 12 p.p. increase over the baseline in the LORO experiment for corn. 
This underscores the superiority of ensemble methods, which are consistently more robust to distribution shifts.
Interestingly, including additional inductive bias in the form of spatial locality (3D-LSTM) further increases the gap between the baseline and the ensemble approach. In the LORO scenario, an improvement of 29 p.p. is achieved for corn, resulting in an overall $R^2$ of 0.76. This score is almost equal to the $ R^2$ score of the baseline model in the standard CV scenario. Similarly, in the LOYO setting, an improvement of 17 p.p. in the $R^2$-score is observed compared to the baseline model. This improvement is proportionally higher than the standard CV experiment (see Tab. \ref{tab:benchmark}). 
\subsection{Distribution shift \& Weight Space Diversity}
To explore the degrading performance and the difference between probabilistic and deterministic models, we analyze the model weights in Fig. \ref{fig:TSNE_Parameters}. The plot illustrates a low-dimensional embedding of the trained model weights of the \gls{de} model using a t-SNE. The weights are displayed for the standard 10-fold CV (left), LOYO (center), and LORO (right) scenarios. Additionally, the weights are colored by folds. For example, in the LOYO scenario, a model is colored by the year in the validation set. The plot reveals interesting insights. While the weight distributions of the ensemble members overlap entirely in the standard 10-fold CV, a clear separation between model parameters is observed under distribution shift (LOYO and LORO). Each fold forms a distinct cluster in weight space with no overlapping. We conclude two main things from this. First, we argue that this may explain the poor performance under data shift, as the model is less capable of generalizing to unknown data distributions due to the separation in weight space. 
Secondly, \glspl{de} explore multiple modes in weight space that may explain the better performance compared to the deterministic baseline, which only explores single modes \cite{fort2019deep}. This is underlined by Fig. \ref{fig:parameter_space_diversity}. The plot illustrates the trajectory in weight space during training. The left row shows the cosine similarity between two ensemble members over the training epochs. Each comparison shows that the similarity in weight space is high at the start of the training and decreases throughout the training. At the end of the training, the weight space is clearly separated between the two ensemble members. 
The right plot shows the trajectories in weight space for two ensemble members, along with the validation loss. The plot underlines that the ensemble members are initialized randomly in nearby regions. Additionally, the plot shows that each ensemble member explores distinct modes in weight space that are characterized by lower loss values. This indicates that each ensemble member explores several optimal solutions, explaining the higher robustness under distribution shifts. This is underlined by \cite{fort2019deep}. More examples are given in the Appendix (see \ref{sec:further_results}).
Although \glspl{de} provide a notable advantage over deterministic models by capturing multiple modes in weight space, their performance still degrades under distribution shift. To address this limitation, we explore the incorporation of additional functional forms into the model. In particular, we employ a 3D-LSTM architecture that captures spatial correlation within the image, a crucial aspect to capture in-field dynamics. The resulting weight space is illustrated in Fig. \ref{fig:TSNE_Parameters_iML}.
As shown in the figure, including more expressive functional forms of the model anchors the solution within closer regions of the weight space. Consequently, generalization under distribution shift improves, as evidenced in Tab. \ref{tab:loro_lofo}, where the distributions exhibit greater overlap, similar to Fig. \ref{fig:TSNE_Parameters} (10-fold cv). Nonetheless, elongated clusters indicate residual uncertainty. 
\section{Conclusion \& Open Challenges}
This work introduced \textit{YieldSAT}, a multimodal dataset for crop yield prediction at both the field and subfield levels. The dataset covers multiple countries, crop types, and years.
We provided benchmark results across several model architectures and highlighted open challenges, including distribution shifts in the ground truth data. Although the dataset enables scalable yield prediction, it does not provide global coverage. For this, more data is required. \\
Several promising research directions remain open. First, most existing approaches operate at the pixel level due to significant variations in field size and the limited data. Modeling entire fields independent of their spatial extent remains an open challenge. 
Second, most models are designed for specific regions or crop types. Recently, Foundation Models (FMs) have emerged as a promising direction, offering the potential to handle multiple \gls{eo} tasks and diverse data sources within a single unified model \cite{guo2024skysense, xiao2025foundation}. Despite their promise, FMs are still in their early stages, especially for regression tasks \cite{xue2025regression}. 
Moreover, integrating uncertainty quantification, physical consistency, and explainability is often overlooked in the current literature but is highly required \cite{zhu2024foundations,najjar2025explainability,najjar2025intrinsic}. The current
analysis is mainly performance-centric, leaving more detailed physical analyses to future work.
Finally, although this study addressed the impact of distribution shifts arising in real-world settings, more attention must be paid to this topic. 
Addressing these challenges is fundamental to improving explainability, fostering trust, and ultimately enabling broader adoption of \gls{dl} to support and advance digital farming.

\section*{Acknowledgments}
This work was partly funded through the ESA InCubed Programme (\url{https://incubed.esa.int/}) as part of the project AI4EO Solution Factory (\url{https://www.ai4eo-solution-factory.de/}). M.M., H.N., and F.M. acknowledge support through a scholarship from the RPTU University of Kaiserslautern-Landau.

{
    \small
    \bibliographystyle{plain}
    \bibliography{main}
}
\clearpage
\appendix
\section{Appendix for YieldSAT: A Multimodal Benchmark Dataset for High-Resolution Crop Yield Prediction}\label{sec:appendix}
\subsection{Crop Values for Data Preprocessing}
The maximum accepted yield values for each crop type are given in Tab. \ref{tab:yield_threshold}. Yield values above this threshold were removed. Additionally, the standard moisture is given in Tab. \ref{tab:yield_threshold} for each crop type that is used to calculate the scaled yield as shown in the main paper. 
\begin{table}[!b]
    \centering
    \caption{Maximum accepted yield values in t/ha for every crop type and the standard moisture.}
    \resizebox{\columnwidth}{!}{
    \begin{tabular}{lcccc}
        \hline
      Standard Values &  Wheat & Rapeseed & Soybean & Corn \\ \hline
      Max. Yield (t/ha) &  20 & 10  & 15  & 45  \\ 
      Standard Moisture (\%) & 15 & 9 & 15 & 16 \\  
        \hline
    \end{tabular}
    }
    \label{tab:yield_threshold}
\end{table}

\subsection{Data Analysis}

\subsubsection{Data Quality and Support Size}\label{sec:quality_support}
\begin{figure}[!b]
    \centering
    \includegraphics[width=0.95\linewidth]{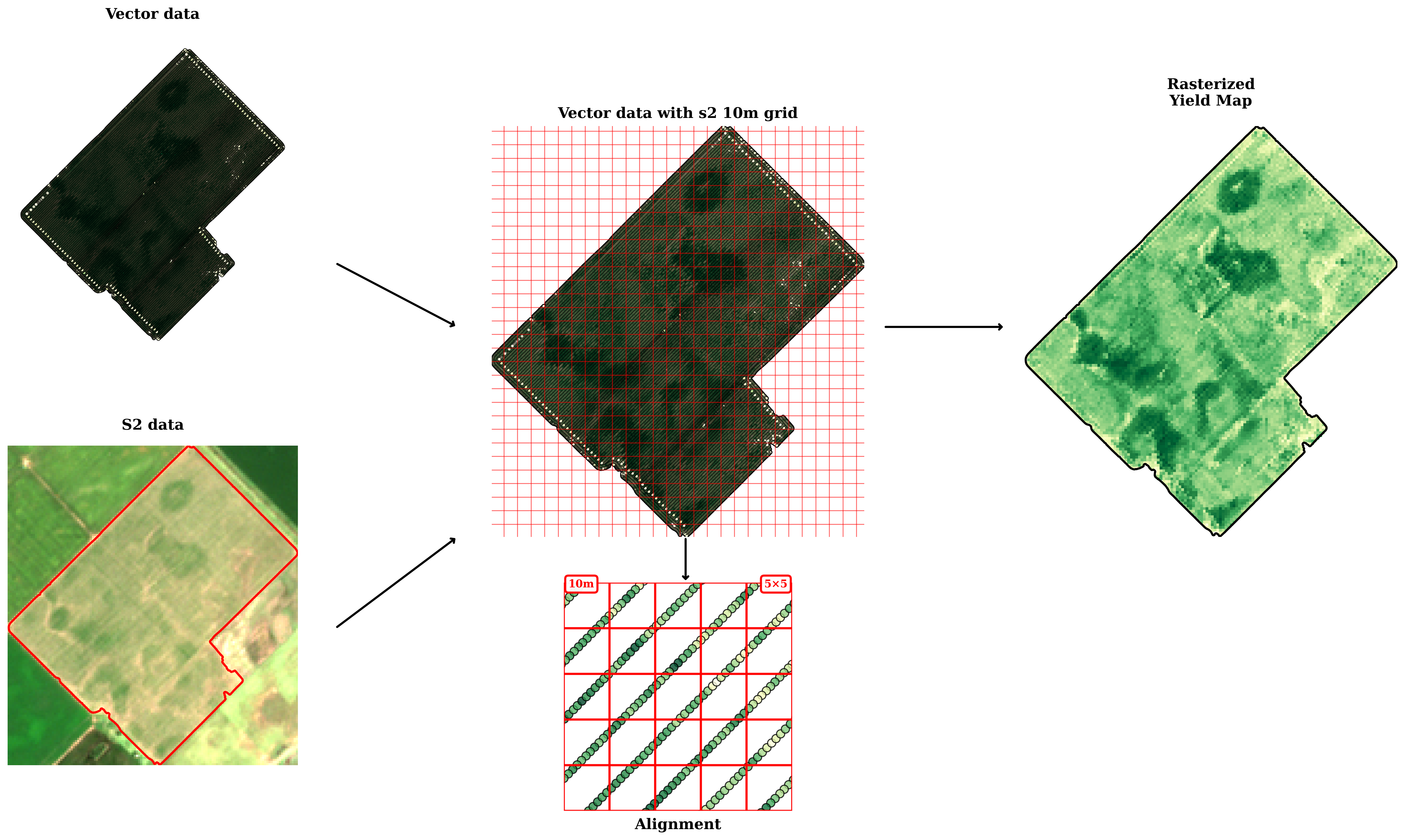}
    \caption{Schematic overview of data collection and preprocessing: A combine harvester collects point vector data of the yield, which is preprocessed and rasterized to align with \gls{s2} $\SI{10}{m}$ grid for pixel-wise regression. }
    \label{fig:rasterization_workflow}
\end{figure}
\begin{figure}[!tb]
    \centering
    \includegraphics[width=.95\linewidth]{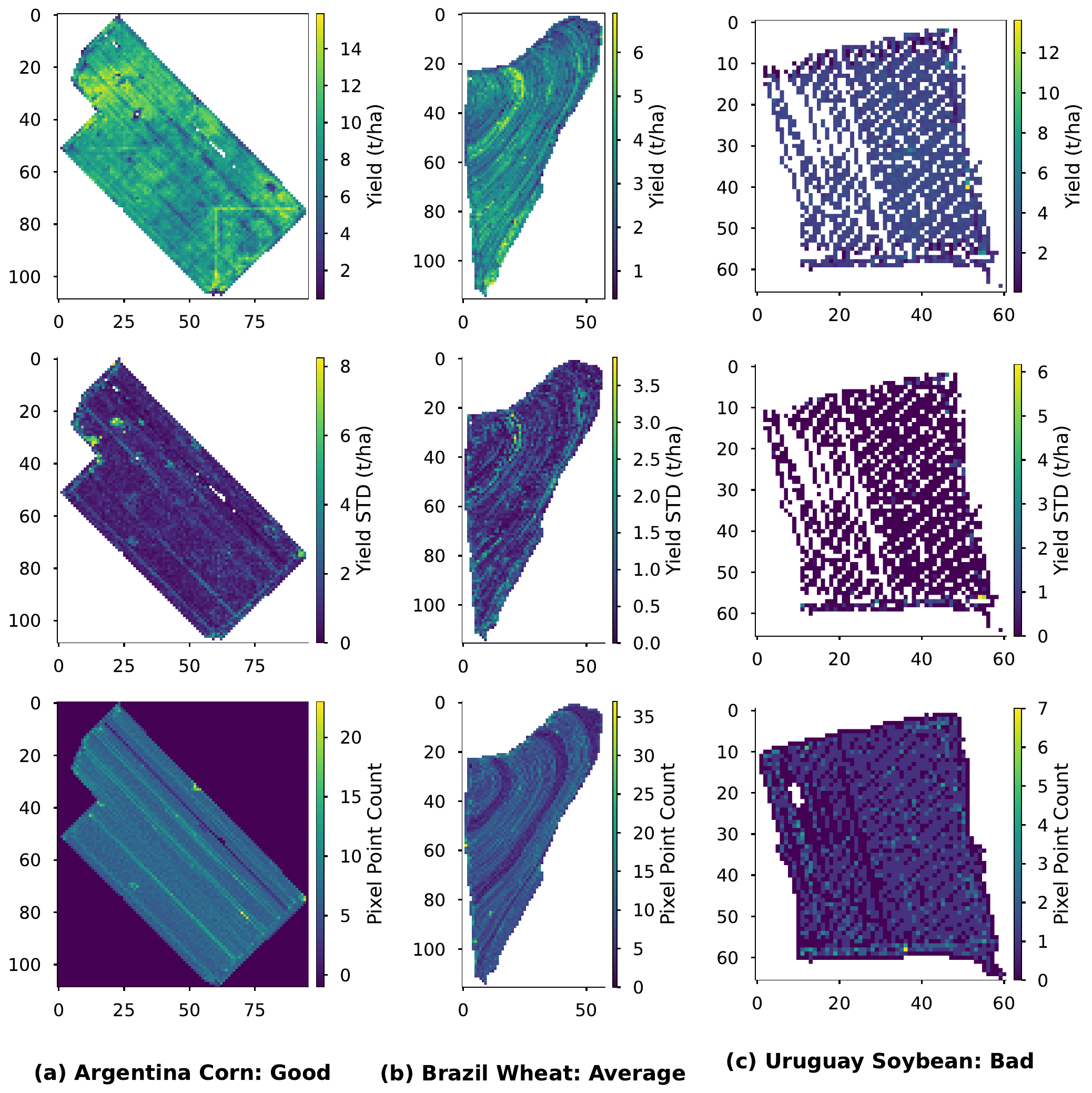}
    \caption{Label support size and standard deviation per $\SI{10}{m}$ pixel caused by the rasterization. Top row: sample count, bottom row: standard deviation. }
    \label{fig:support_sizes}
\end{figure}

\begin{figure}[!t]
    \centering
    \includegraphics[width=.95\linewidth]{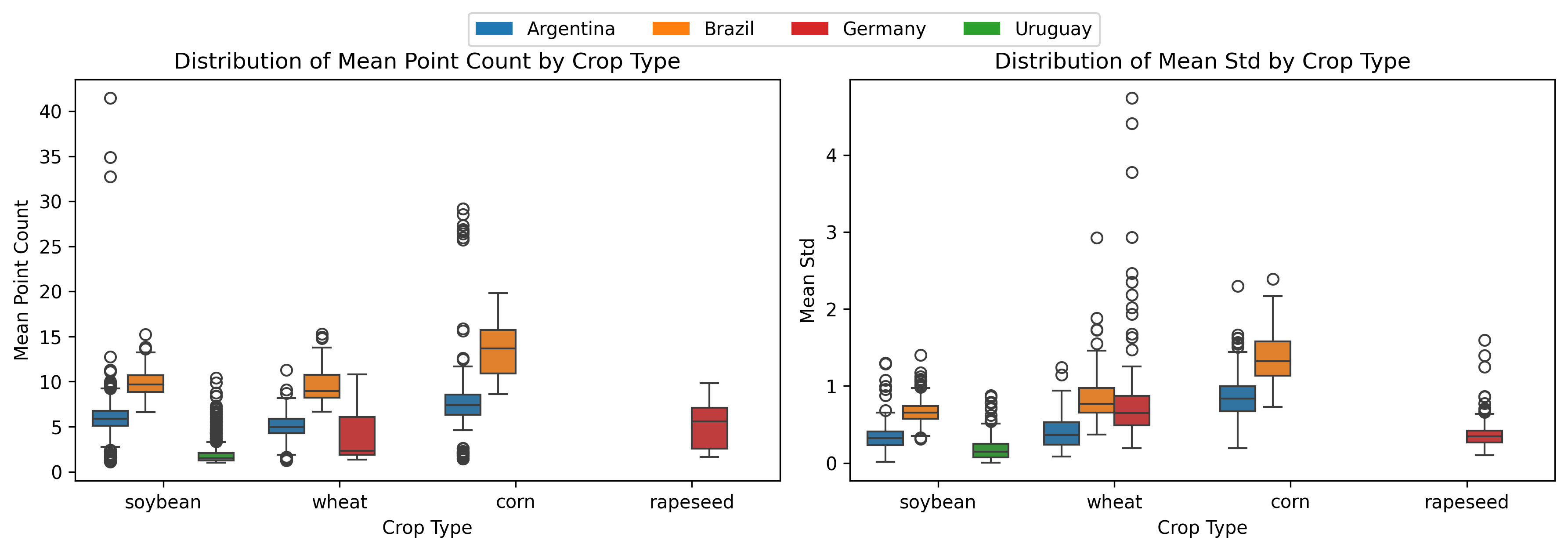}
    \caption{Boxplots for pixel point count (left) and mean standard deviation (right) per crop and country.}
    \label{fig:support_sizes_dataset}
\end{figure}
The rasterization workflow for a single field is shown in Fig. \ref{fig:rasterization_workflow}. Combine harvester in vector format must be processed to match high-resolution \gls{eo} inputs (e.g., \gls{s2}). For this, every yield measurement that falls into a grid cell of $\SI{10}{m} \times \SI{10}{m}$ are averaged. Due to harvester path density, swath width, speed, logging frequency, and positional delay, different numbers of yield measurements can fall into a single grid cell, resulting in variable support sizes for every pixel. Those support sizes are often correlated in space. An example of different support sizes for every yield map quality level, shown in Fig. \ref{fig:yield_map_quality}, is illustrated in Fig. \ref{fig:support_sizes}. The figure shows the number of points that fall into a grid cell and the corresponding standard deviation per pixel. It is evident that the yield maps have different support sizes and variability per pixel. In Fig. \ref{fig:support_sizes_dataset}, we show that these characteristics depend on the crop type and dataset. For instance, Brazil has consistently more yield points per pixel, while Uruguay has fewer samples. This is also reflected in a lower standard deviation per pixel. For each yield map, we also provide the raster grid for the variable support size. \\
Each field was manually inspected and curated. For this, each field is assigned a quality score of either "good", "average", or "bad", following a stringent guideline:\\
\textbf{1. Good} yield maps are characterized by high data quality and consistency. They do not exhibit strong visible striping artifacts (small can be present). The data coverage is dense, with no significant gaps or sparsity. Additionally, good yield maps show meaningful variability across the field, reflecting realistic yield differences, while maintaining smooth spatial transitions without abrupt or isolated changes. \\
\textbf{2. Bad} yield maps display clear quality issues. These may include visible striping patterns, often caused by sensor or machinery inconsistencies, as well as sparse or incomplete data coverage. Such maps typically lack meaningful variability, with large areas showing nearly constant values. They may also contain isolated pixels or small regions with sudden, unrealistic changes in yield values that are inconsistent with the surrounding data. \\
\textbf{3. Average} yield maps fall between these two extremes. They may exhibit minor striping, moderate sparsity, or limited variability, but not to the extent observed in bad maps. While generally usable, they do not meet the quality standards of good yield maps and may require additional processing or careful interpretation. \\
An example of each class is shown in the main paper in Fig. \ref{fig:yield_map_quality}.
\begin{figure}[!tb]
    \centering
    \includegraphics[width=0.85\linewidth]{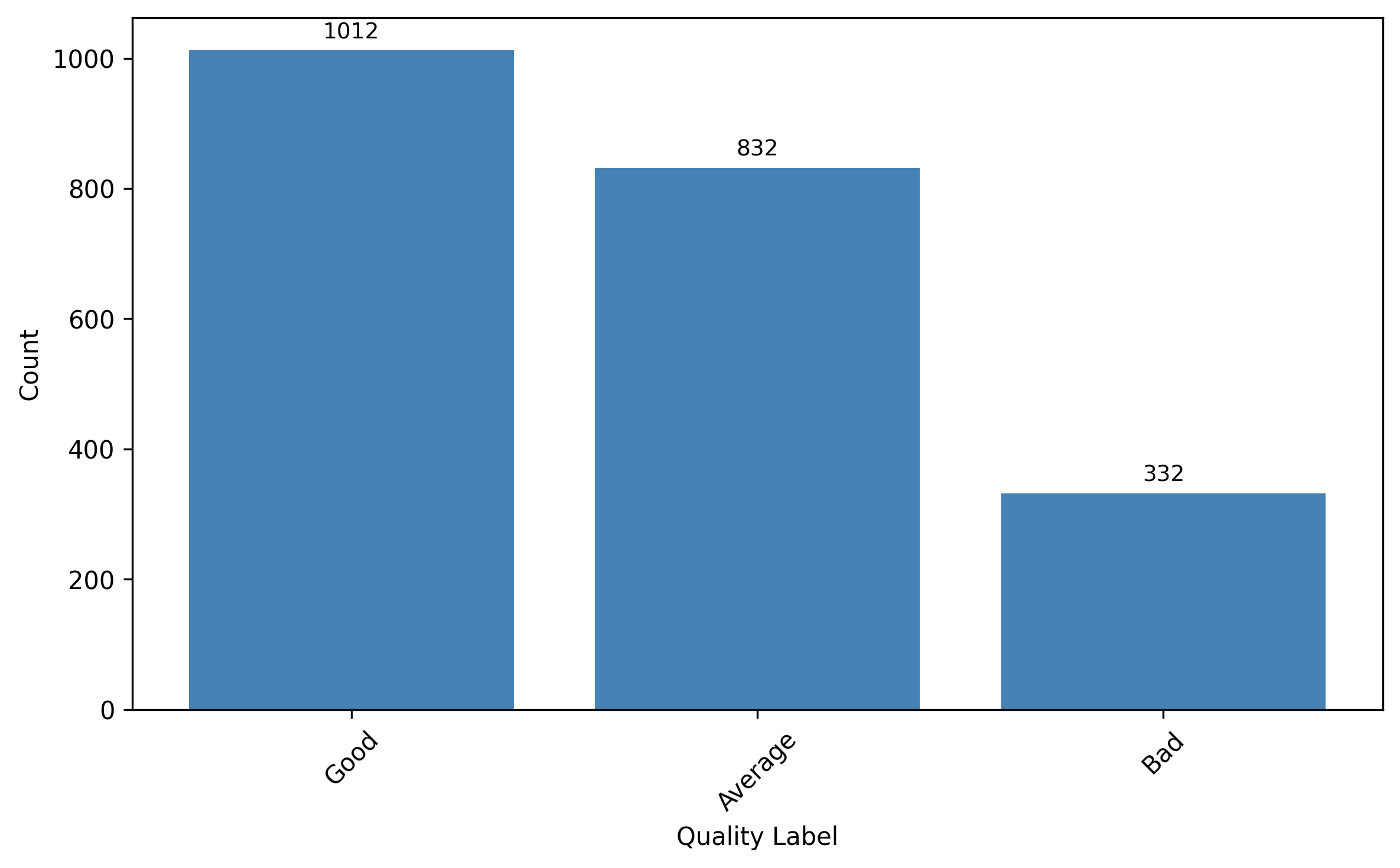}
    \caption{Overall distribution of yield map quality over the entire dataset.}
    \label{fig:overall_quality}
\end{figure}
\begin{figure}[!tb]
    \centering
    \includegraphics[width=0.85\linewidth]{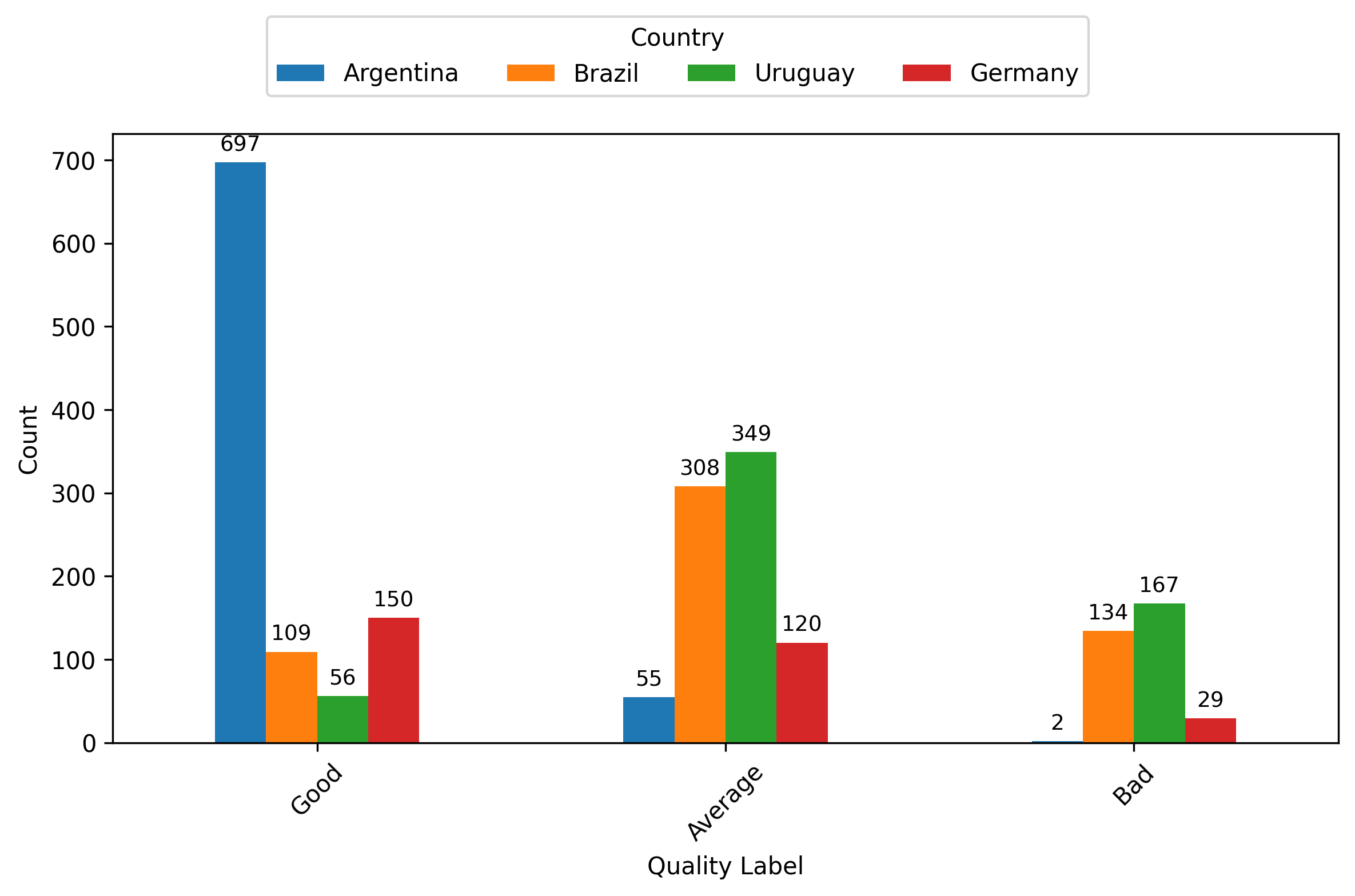}
    \caption{Overall distribution of yield map quality over the entire dataset grouped by country. }
    \label{fig:overall_quality_by country}
\end{figure}
\begin{figure}[!tb]
    \centering
    \includegraphics[width=0.85\linewidth]{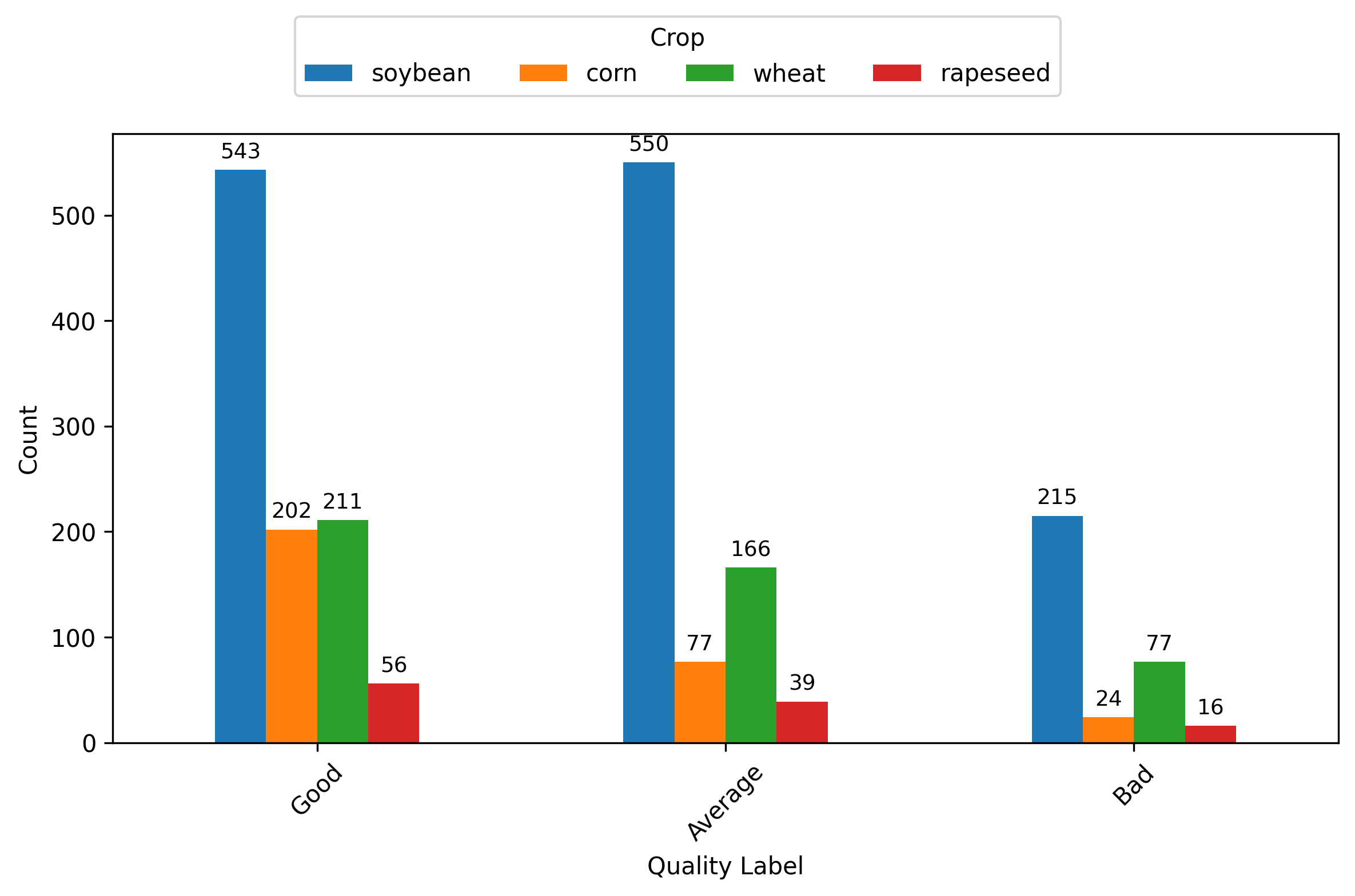}
    \caption{Overall distribution of yield map quality over the entire dataset grouped by country. }
    \label{fig:overall_quality_by crop}
\end{figure}
The distribution of yield map quality across the entire dataset is shown in Fig. \ref{fig:overall_quality}. The plot indicates that the majority of yield maps are of good quality, followed by average and bad categories, based on a total of 332 fields.
Fig. \ref{fig:overall_quality_by country} presents the distribution of yield map quality grouped by country. Argentina exhibits the highest number of good-quality yield maps, whereas Uruguay has the largest number of low-quality (bad) yield maps.
Finally, Fig. \ref{fig:overall_quality_by crop} shows the distribution of yield map quality by crop type. Soybean accounts for the largest number of high-quality yield maps, followed by wheat and corn. \\
The quality labels are stored in the metadata of each field. 

\subsubsection{Distribution Shifts: Countries and Crop Types}\label{sec:distribution_shifts_country_crop}
To test whether the distributions of crop types and countries are significantly different, significance testing is performed to compare the distributions between groups. Both tests demonstrate significantly different distributions between countries and crop types (p-values $<0.0001$) as shown in Tab. \ref{tab:kruskals_walliis_crountry_crop}.
Following, we perform a post-hoc analysis that compares pairs within each group using Dunn’s significance test using a Holm-Bonferroni correction. The results for the pairwise comparison between each country are shown in Tab. \ref{tab:dunns_country}. Notably, each group has a distinct yield distribution, with most p-values $<0.0001$. Except for between Germany and Brazil, no significance is present. Additionally, the Dunn’s test for the pairwise comparison between crop types is shown in Tab. \ref{table:dunns_crop}. Similarly, most distributions differ significantly between crop types, with all p-values $<0.0001$, except for rapeseed and soybean.

\subsubsection{Distribution Shifts: Regions and Years} \label{sec:distribution_shifts_year_region}
\begin{figure}[!tb]
    \centering
    \includegraphics[width=0.95\linewidth]{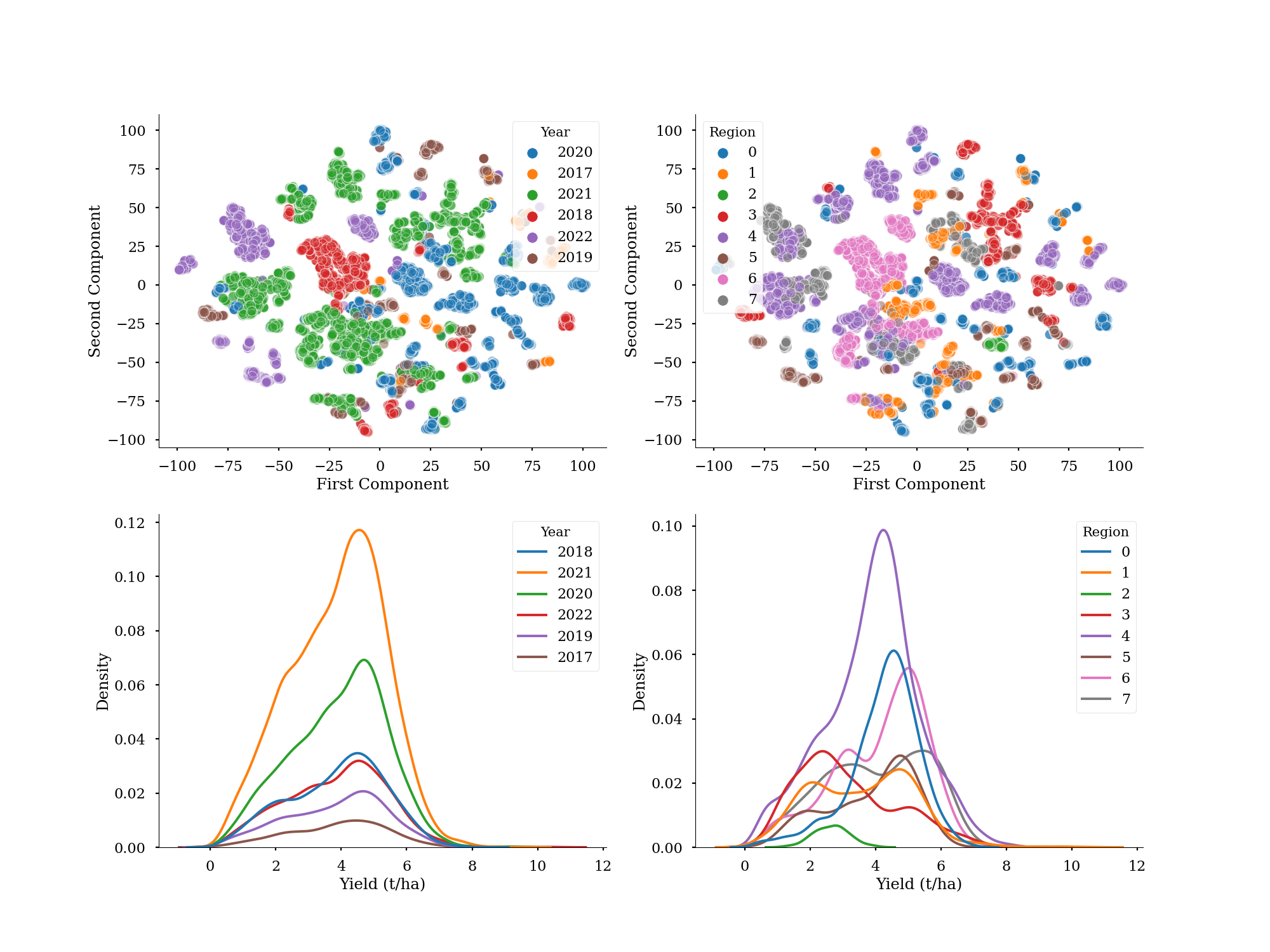}
    \caption{Visualization of the surface reflectance and yield data distribution only for soybean in Argentina, grouped by years and regions. Top left: t-SNE plot of the S2 surface reflectance data grouped by \textbf{years}. Top right: t-SNE plot of the S2 surface reflectance data grouped by \textbf{region}. Bottom left: Kernel density estimation plot for the target yield data grouped by \textbf{years}. Bottom right: Kernel density estimation plot for the target yield data grouped by \textbf{regions}. }
    \label{fig:TSNE_S2_MX_year_region}
\end{figure}

Importantly, we also find significant differences in data distribution and surface reflectance across regions and years within a single country and crop type. In Fig. \ref{fig:TSNE_S2_MX_year_region} t-SNE plot of the surface reflectance for only soybean in Argentina is shown. The data are colored by year (left) and region (right). We highlight that, in both cases, a separation between individual years and regions is evident, as shown by clusters within each group. 
Below is the kernel density estimation plot of the target yield distribution. The data are grouped by year (left) and region (right). Note that each group's distribution is unique. Individual distributions do not follow a normal distribution. Moreover, the plot shows that the yield distribution differs between years and regions. The results indicate unique patterns within each year-region combination across countries and crop types. 
\begin{table}[!tb]
    \centering
    \caption{\label{table:kruskals_wallis}
     Kruskal–Wallis H-test between yield distributions grouped by countries and by
    crop type. $**** = p < 0.0001$.
    }
    \resizebox{.8\columnwidth}{!}{
        \begin{tabular}{l|cc}
\hline
\multicolumn{1}{c|}{\textbf{Evaluation}} & \textbf{Country}           & \textbf{Crop}         \\ \hline
p-value                                  &  **** &  **** \\ \hline
\end{tabular}
    } 
    \label{tab:kruskals_walliis_crountry_crop}
\end{table}
\begin{table}[!tb]
    \centering
    \caption{
        Pairwise post-hoc comparisons of the yield distributions between individual \textbf{countries}. Each cell displays the statistical significance level of the difference between two countries based on the Dunn’s test using the Holm–Bonferroni correction. ns = no significance ($p \geq 0.05$), $* = p < 0.05$, $** = p < 0.01$, $*** = p < 0.001$, $**** = p < 0.0001$.  
    }
    \resizebox{\columnwidth}{!}{
        \begin{tabular}{lcccc}
\hline
\multicolumn{1}{l|}{\textbf{Comparision}} &
  \multicolumn{1}{c}{\textbf{Argentina}} &
  \multicolumn{1}{c}{\textbf{Brazil}} &
  \multicolumn{1}{c}{\textbf{Germany}} &
  \multicolumn{1}{c}{\textbf{Uruguay}} \\ \hline
\multicolumn{1}{l|}{\textbf{Argentina}} & -   & **** & **** & **** \\
\multicolumn{1}{l|}{\textbf{Brazil}}    & **** & -   & ns   & **** \\
\multicolumn{1}{l|}{\textbf{Germany}}   & **** & ns   & -   & **** \\
\multicolumn{1}{l|}{\textbf{Uruguay}}   & **** & **** & **** & -   \\ \hline
\end{tabular}

    } 
    \label{tab:dunns_country}
\end{table}
\begin{table}[!tb]
    \centering
    \caption{\label{table:dunns_crop}
        Pairwise post-hoc comparisons of the yield distributions between individual \textbf{crops}. Each cell displays the statistical significance level of the difference between two crops based on the Dunn’s test using the Holm–Bonferroni correction. ns = no significance ($p \geq 0.05$), $* = p < 0.05$, $** = p < 0.01$, $*** = p < 0.001$, $**** = p < 0.0001$.     }
    \resizebox{\columnwidth}{!}{
        \begin{tabular}{l|cccc}
\hline
\multicolumn{1}{l|}{\textbf{Comparision}} &
  \multicolumn{1}{c}{\textbf{Corn}} &
  \multicolumn{1}{c}{\textbf{Rapeseed}} &
  \multicolumn{1}{c}{\textbf{Soybean}} &
  \multicolumn{1}{c}{\textbf{Wheat}} \\ \hline
\textbf{Corn}     & -   & **** & **** & **** \\
\textbf{Rapeseed} & **** & -   & ns   & **** \\
\textbf{Soybean}  & **** & ns   & -   & **** \\
\textbf{Wheat}    & **** & **** & **** & -   \\ \hline
\end{tabular}
    } 
    \label{tab:dunns_crops}
\end{table}
\begin{table}[!tb]
    \centering
    \caption{Kruskal–Wallis H-test between yield distributions for soybean in Argentina, grouped by years and by regions. $**** = p < 0.0001$. 
    }
    \resizebox{.8\columnwidth}{!}{
        \begin{tabular}{l|cc}
\hline
\multicolumn{1}{c|}{\textbf{Evaluation}} & \textbf{Year}           & \textbf{Region}         \\ \hline
p-value                                  & **** &  **** \\ \hline
\end{tabular}
    } 
    \label{tab:kurskals_years_region}
\end{table}
\begin{table}[!tb]
    \centering
    \caption{
    Pairwise post-hoc comparisons of the yield distributions for soybean in Argentina between individual \textbf{years}. Each cell displays the statistical significance level of the difference between two years based on Dunn’s test using the Holm–Bonferroni correction. ns = no significance ($p \geq 0.05$), $* = p < 0.05$, $** = p < 0.01$, $*** = p < 0.001$, $**** = p < 0.0001$. 
    }
    \resizebox{\columnwidth}{!}{
        \begin{tabular}{c|cccccc}
\hline
\multicolumn{1}{l|}{\textbf{Year}} & \textbf{2017} & \textbf{2018} & \textbf{2019} & \textbf{2020} & \textbf{2021} & \textbf{2022} \\ \hline
\textbf{2017} & -    & **** & ***  & **** & **** & **** \\
\textbf{2018} & **** & -    & ns   & ***  & **** & **** \\
\textbf{2019} & ***  & ns   & -    & **   & **** & **** \\
\textbf{2020} & **** & ***  & **   & -    & **** & **** \\
\textbf{2021} & **** & **** & **** & **** & -    & **** \\
\textbf{2022} & **** & **** & **** & **** & **** & -    \\ \hline
\end{tabular}

    } 
    \label{tab:t_test_year}
\end{table}
\begin{table}[!tb]
    \centering
    \caption{
    Pairwise post-hoc comparisons of the yield distributions for soybean in Argentina between individual \textbf{regions}. Each cell displays the statistical significance level of the difference between two regions based on Dunn’s test using the Holm–Bonferroni correction. ns = no significance ($p \geq 0.05$), $* = p < 0.05$, $** = p < 0.01$, $*** = p < 0.001$, $**** = p < 0.0001$. 
    }
    \resizebox{\columnwidth}{!}{
        \begin{tabular}{c|cccccccc}
\hline
\textbf{Region} & \textbf{0} & \textbf{1} & \textbf{2} & \textbf{3} & \textbf{4} & \textbf{5} & \textbf{6} & \textbf{7} \\ \hline
\textbf{0} & -   & **** & **** & **** & **** & **** & **   & **** \\
\textbf{1} & **** & -   & **** & **** & **** & **** & **** & **** \\
\textbf{2} & **** & **** & -   & ***  & **** & **** & **** & **** \\
\textbf{3} & **** & **** & ***  & -   & **** & **** & **** & **** \\
\textbf{4} & **** & **** & **** & **** & -   & ns   & **** & ns   \\
\textbf{5} & **** & **** & **** & **** & ns   & -   & ns   & ns   \\
\textbf{6} & **   & **** & **** & **** & **** & ns   & -   & ***  \\
\textbf{7} & **** & **** & **** & **** & ns   & ns   & ***  & -   \\ \hline
\end{tabular}

    } 
    \label{tab:t_test_region}
\end{table}
To test the hypothesis that all groups have the same yield distribution, a Kruskal-Wallis test is performed to compare the distributions across groups. The results are depicted in Tab. \ref{tab:kurskals_years_region}. For both tests, a significantly high p-value is reported ($p<0.0001$). This means that at least one year and one region have a significantly different yield distribution compared to the rest of the groups.  
To compare each pair of yield distributions, a post hoc test is performed using Dunn's test. 
The pairwise comparison between the years is depicted in Tab. \ref{tab:t_test_year}. We highlight that most pairwise comparisons indicate significantly different distributions. Only between 2018 and 2019 is there no significance. 
The pairwise comparison between the regions is depicted in Tab. \ref{tab:t_test_region}. As in previous years, the data distribution across regions is mostly significantly different. However, individual regions do not show a significantly different distribution. For instance, region 5 shows no significant difference between regions 4, 6, and 7.\\

In conclusion, the results undermine the assumption that the distribution across years and regions is mostly significantly different, which increases the difficulty of generalization.

\begin{figure}[!htb]
  \centering
  \includegraphics[width=.95\columnwidth]{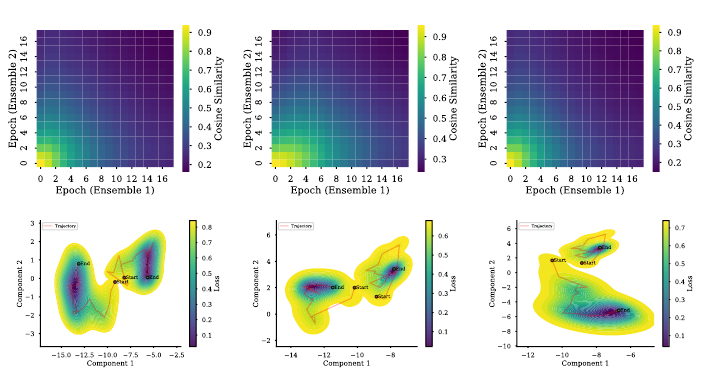}
  \caption{Visualization of the weight space diversity during model training. Top: Cosine similarity between pairs of ensemble members during training. Bottom: PCA plot of the weight space during training, together with the loss. The trajectory in the weight space is highlighted in red from start to end.  }
\label{fig:parameter_space_diversity_apdx}
\end{figure}

\subsection{Further Experiments}\label{sec:further_results}
\begin{figure*}[!tb]
  \centering
  \includegraphics[width=.95\linewidth]{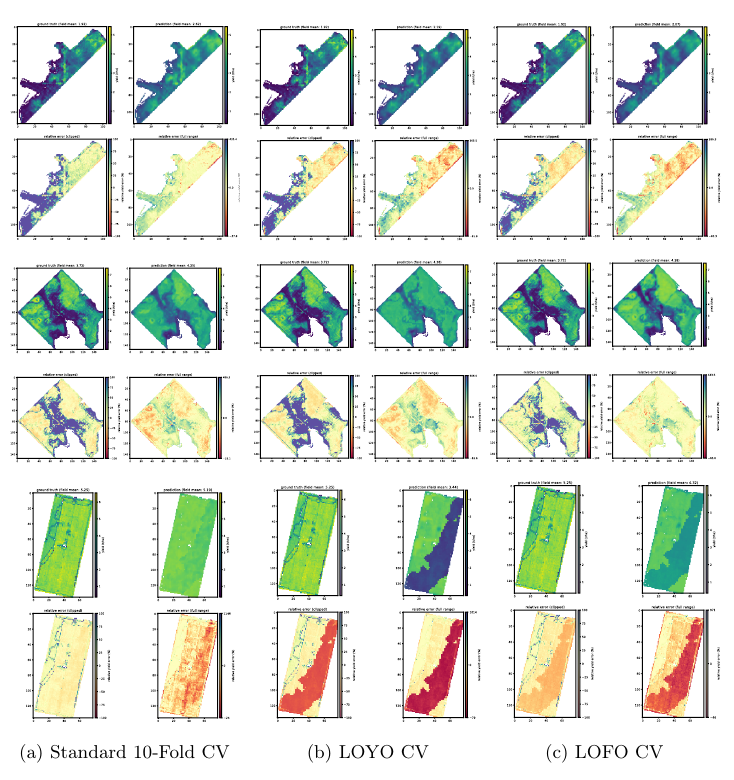}
  \caption{Example ground truth and predicted yield maps together with the pixel-wise error for the standard CV, the LORO, and LOYO CV under distribution shift. The results are generated with the LSTM model \cite{pathak}, trained on \gls{s2} data.}
\label{fig:example}
\end{figure*}

Fig. \ref{fig:example} shows additional example predictions for single fields (image), illustrating the ground-truth yield map, the predicted yield map, and the pixel-wise error (clipped and full range). We show the results for the standard 10-fold cv experiment and the temporal (LOYO) and spatial (LORO) experiments. Note that for some fields, the LOYO and LORO experiments exhibit severe performance collapse, as evidenced by high pixel-wise errors and areas of mode collapse (only single-scalar predictions with no spatial variability). \\
In Tab. \ref{tab:full_replication_cv10_subfield_field} we provide a full replication for the  10-fold cv baseline experiment of all benchmark models for each dataset at the field level and at the subfield (pixel)-level in Tab. \ref{tab:full_replication_cv10_subfield_field}. We further provide a full replication of all benchmark models on the OOD experiments, namely the temporal (LOYO) and spatial (LORO) splitting. For the temporal splitting, the field level results are given in Tab. \ref{tab:full_replication_loyo_field} and for the subfield (pixel)-level in Tab. \ref{tab:table_all_loyo_subfield}. For the spatial splitting (LORO), the field level results are given in Tab. \ref{tab:table_all_loro_field_leve} and for the subfield (pixel)-level in Tab. \ref{tab:table_all_loro_subfield}.  

Fig. \ref{fig:parameter_space_diversity_apdx} provides more examples of the cosine similarity in weight space over the training epochs, together with the trajectory in weight space. 

\begin{table*}[!tb]
    \centering
    \caption{
    Results for the RMSE (t/ha) ($\downarrow$) and the $R^2$-score ($\uparrow$) for different models and datasets for the standard 10-fold cross-validation at the field level. ARG = Argentina, BRA = Brazil, GER = Germany, URG = Uruguay. C = corn, R = rapeseed, S = soybean, W = wheat.
    }
    \resizebox{.95\linewidth}{!}{
\begin{tabular}{ccclllllllllllllllllll}
\hline
\multicolumn{4}{c}{\multirow{2}{*}{\textbf{Evaluation}}} &
  \multicolumn{18}{c}{\textbf{Field-Level}} \\ \cline{5-22} 
\multicolumn{4}{c}{} &
  \multicolumn{2}{c}{\textbf{ARG-C}} &
  \multicolumn{2}{c}{\textbf{ARG-S}} &
  \multicolumn{2}{c}{\textbf{ARG-W}} &
  \multicolumn{2}{c}{\textbf{BRA-C}} &
  \multicolumn{2}{c}{\textbf{BRA-S}} &
  \multicolumn{2}{c}{\textbf{BRA-W}} &
  \multicolumn{2}{c}{\textbf{GER-R}} &
  \multicolumn{2}{c}{\textbf{GER-W}} &
  \multicolumn{2}{c}{\textbf{URG-S}} \\ \hline
\textbf{Splitting} &
  \textbf{Modalities} &
  \textbf{Fusion Method} &
  \multicolumn{1}{c}{\textbf{Model}} &
  $R^2$ &
  RMSE &
  $R^2$ &
  RMSE &
  $R^2$ &
  RMSE &
  $R^2$ &
  RMSE &
  $R^2$ &
  RMSE &
  $R^2$ &
  RMSE &
  $R^2$ &
  RMSE &
  $R^2$ &
  RMSE &
  $R^2$ &
  RMSE \\ \hline
\multirow{10}{*}{\textbf{CV10}} &
  \multirow{4}{*}{Sentinel-2} &
  \multicolumn{1}{c}{X} &
  3D-ConvLSTM &
  0.84 {\tiny $\pm$ \textcolor{black!50}{0.07}} &
  1.13 {\tiny $\pm$ \textcolor{black!50}{0.22}} &
  0.79 {\tiny $\pm$ \textcolor{black!50}{0.12}} &
  0.55 {\tiny $\pm$ \textcolor{black!50}{0.16}} &
  0.92 {\tiny $\pm$ \textcolor{black!50}{0.08}} &
  0.62 {\tiny $\pm$ \textcolor{black!50}{0.24}} &
  0.82 {\tiny $\pm$ \textcolor{black!50}{0.09}} &
  0.74 {\tiny $\pm$ \textcolor{black!50}{0.12}} &
  0.76 {\tiny $\pm$ \textcolor{black!50}{0.09}} &
  0.34 {\tiny $\pm$ \textcolor{black!50}{0.05}} &
  0.73 {\tiny $\pm$ \textcolor{black!50}{0.09}} &
  0.42 {\tiny $\pm$ \textcolor{black!50}{0.16}} &
  0.81 {\tiny $\pm$ \textcolor{black!50}{0.10}} &
  0.58 {\tiny $\pm$ \textcolor{black!50}{0.19}} &
  0.65 {\tiny $\pm$ \textcolor{black!50}{0.16}} &
  1.12 {\tiny $\pm$ \textcolor{black!50}{0.34}} &
  0.77 {\tiny $\pm$ \textcolor{black!50}{0.06}} &
  0.51 {\tiny $\pm$ \textcolor{black!50}{0.10}} \\
\hline
\end{tabular}

    } 
    \label{tab:full_replication_cv10_subfield_field}
\end{table*}

\begin{table*}[!tb]
    \centering
    \caption{
    Results for the RMSE (t/ha) ($\downarrow$) and the $R^2$-score ($\uparrow$) for different models and datasets for the standard 10-fold cross-validation at the subfield (pixel)-level. ARG = Argentina, URG = Uruguay, GER = Germany, BRA = Brazil. S = soybean, R = rapeseed, W = wheat, C = Corn.
    }
    \resizebox{.95\linewidth}{!}{
\begin{tabular}{ccclllllllllllllllllll}
\hline
\multicolumn{4}{c}{\multirow{2}{*}{\textbf{Evaluation}}} &
  \multicolumn{18}{c}{\textbf{Subfield   (Pixel)-Level}} \\ \cline{5-22} 
\multicolumn{4}{c}{} &
  \multicolumn{2}{c}{\textbf{ARG-C}} &
  \multicolumn{2}{c}{\textbf{ARG-S}} &
  \multicolumn{2}{c}{\textbf{ARG-W}} &
  \multicolumn{2}{c}{\textbf{BRA-C}} &
  \multicolumn{2}{c}{\textbf{BRA-S}} &
  \multicolumn{2}{c}{\textbf{BRA-W}} &
  \multicolumn{2}{c}{\textbf{GER-R}} &
  \multicolumn{2}{c}{\textbf{GER-W}} &
  \multicolumn{2}{c}{\textbf{URG-S}} \\ \hline
\textbf{Splitting} &
  \textbf{Modalities} &
  \textbf{Fusion Method} &
  \multicolumn{1}{c}{\textbf{Model}} &
  R2 &
  RMSE &
  R2 &
  RMSE &
  R2 &
  RMSE &
  R2 &
  RMSE &
  R2 &
  RMSE &
  R2 &
  RMSE &
  R2 &
  RMSE &
  R2 &
  RMSE &
  R2 &
  RMSE \\ \hline
\multirow{10}{*}{\textbf{CV10}} &
  \multirow{4}{*}{Sentinel-2} &
  \multicolumn{1}{l}{X} &
  3D-ConvLSTM &
  0.65 {\tiny $\pm$ \textcolor{black!50}{0.11}} &
  2.2 {\tiny $\pm$ \textcolor{black!50}{0.27}} &
  0.65 {\tiny $\pm$ \textcolor{black!50}{0.06}} &
  0.9 {\tiny $\pm$ \textcolor{black!50}{0.11}} &
  0.79 {\tiny $\pm$ \textcolor{black!50}{0.1}} &
  1.1 {\tiny $\pm$ \textcolor{black!50}{0.18}} &
  0.45 {\tiny $\pm$ \textcolor{black!50}{0.12}} &
  2.13 {\tiny $\pm$ \textcolor{black!50}{0.31}} &
  0.39 {\tiny $\pm$ \textcolor{black!50}{0.07}} &
  0.94 {\tiny $\pm$ \textcolor{black!50}{0.06}} &
  0.23 {\tiny $\pm$ \textcolor{black!50}{0.05}} &
  1.38 {\tiny $\pm$ \textcolor{black!50}{0.18}} &
  0.49 {\tiny $\pm$ \textcolor{black!50}{0.1}} &
  1.2 {\tiny $\pm$ \textcolor{black!50}{0.22}} &
  0.34 {\tiny $\pm$ \textcolor{black!50}{0.09}} &
  2.4 {\tiny $\pm$ \textcolor{black!50}{0.34}} &
  0.41 {\tiny $\pm$ \textcolor{black!50}{0.06}} &
  1.22 {\tiny $\pm$ \textcolor{black!50}{0.12}} \\
\hline
\end{tabular}

    } 
    \label{tab:full_replication_cv10_subfield}
\end{table*}

\begin{table*}[!tb]
    \centering
    \caption{
    Results for the RMSE (t/ha) ($\downarrow$) and the $R^2$-score ($\uparrow$) for different models and datasets for the Leave-One-Year-Out cross-validation at the field level. ARG = Argentina, BRA = Brazil, GER = Germany, URG = Uruguay. C = corn, R = rapeseed, S = soybean, W = wheat.
    }
    \resizebox{.95\linewidth}{!}{
\begin{tabular}{ccclllllllllllllllllll}
\hline
\multicolumn{4}{c}{\multirow{2}{*}{\textbf{Evaluation}}} &
  \multicolumn{18}{c}{\textbf{Field-Level}} \\ \cline{5-22} 
\multicolumn{4}{c}{} &
  \multicolumn{2}{c}{\textbf{ARG-C}} &
  \multicolumn{2}{c}{\textbf{ARG-S}} &
  \multicolumn{2}{c}{\textbf{ARG-W}} &
  \multicolumn{2}{c}{\textbf{BRA-C}} &
  \multicolumn{2}{c}{\textbf{BRA-S}} &
  \multicolumn{2}{c}{\textbf{BRA-W}} &
  \multicolumn{2}{c}{\textbf{GER-R}} &
  \multicolumn{2}{c}{\textbf{GER-W}} &
  \multicolumn{2}{c}{\textbf{URG-S}} \\ \hline
\textbf{Splitting} &
  \textbf{Modalities} &
  \textbf{Fusion Method} &
  \multicolumn{1}{c}{\textbf{Model}} &
  $R^2$ &
  RMSE &
  $R^2$ &
  RMSE &
  $R^2$ &
  RMSE &
  $R^2$ &
  RMSE &
  $R^2$ &
  RMSE &
  $R^2$ &
  RMSE &
  $R^2$ &
  RMSE &
  $R^2$ &
  RMSE &
  $R^2$ &
  RMSE \\ \hline
\multirow{10}{*}{\textbf{LOYO}} &
  \multirow{4}{*}{Sentinel-2} &
  X &
  3D-ConvLSTM &
  0.61 {\tiny $\pm$ \textcolor{black!50}{0.55}} &
  1.75 {\tiny $\pm$ \textcolor{black!50}{0.51}} &
  0.65 {\tiny $\pm$ \textcolor{black!50}{0.31}} &
  0.71 {\tiny $\pm$ \textcolor{black!50}{0.18}} &
  0.78 {\tiny $\pm$ \textcolor{black!50}{0.88}} &
  1.00 {\tiny $\pm$ \textcolor{black!50}{0.40}} &
  0.20 {\tiny $\pm$ \textcolor{black!50}{0.32}} &
  1.58 {\tiny $\pm$ \textcolor{black!50}{0.75}} &
  0.36 {\tiny $\pm$ \textcolor{black!50}{0.42}} &
  0.55 {\tiny $\pm$ \textcolor{black!50}{0.15}} &
  -0.06 {\tiny $\pm$ \textcolor{black!50}{1.30}} &
  0.82 {\tiny $\pm$ \textcolor{black!50}{0.42}} &
  0.43 {\tiny $\pm$ \textcolor{black!50}{0.24}} &
  1.02 {\tiny $\pm$ \textcolor{black!50}{0.23}} &
  0.23 {\tiny $\pm$ \textcolor{black!50}{0.60}} &
  1.67 {\tiny $\pm$ \textcolor{black!50}{0.60}} &
  0.63 {\tiny $\pm$ \textcolor{black!50}{0.25}} &
  0.65 {\tiny $\pm$ \textcolor{black!50}{0.15}} \\
\hline
\end{tabular}

    } 
    \label{tab:full_replication_loyo_field}
\end{table*}
\begin{table*}[!tb]
    \centering
    \caption{
    Results for the RMSE (t/ha) ($\downarrow$) and the $R^2$-score ($\uparrow$) for different models and datasets for the Leave-One-Region-Out (LORO) cross-validation at the field level.ARG = Argentina, BRA = Brazil, GER = Germany, URG = Uruguay. C = corn, R = rapeseed, S = soybean, W = wheat.
    }
    \resizebox{.95\linewidth}{!}{
\begin{tabular}{ccclllllllllllllllllll}
\hline
\multicolumn{4}{c}{\multirow{2}{*}{\textbf{Evaluation}}} &
  \multicolumn{18}{c}{\textbf{Subfield   (Pixel)-Level}} \\ \cline{5-22} 
\multicolumn{4}{c}{} &
  \multicolumn{2}{c}{\textbf{ARG-C}} &
  \multicolumn{2}{c}{\textbf{ARG-S}} &
  \multicolumn{2}{c}{\textbf{ARG-W}} &
  \multicolumn{2}{c}{\textbf{BRA-C}} &
  \multicolumn{2}{c}{\textbf{BRA-S}} &
  \multicolumn{2}{c}{\textbf{BRA-W}} &
  \multicolumn{2}{c}{\textbf{GER-R}} &
  \multicolumn{2}{c}{\textbf{GER-W}} &
  \multicolumn{2}{c}{\textbf{URG-S}} \\ \hline
\textbf{Splitting} &
  \textbf{Modalities} &
  \textbf{Fusion Method} &
  \multicolumn{1}{c}{\textbf{Model}} &
  $R^2$ &
  RMSE &
  $R^2$ &
  RMSE &
  $R^2$ &
  RMSE &
  $R^2$ &
  RMSE &
  $R^2$ &
  RMSE &
  $R^2$ &
  RMSE &
  $R^2$ &
  RMSE &
  $R^2$ &
  RMSE &
  $R^2$ &
  RMSE \\ \hline
\multirow{10}{*}{\textbf{LOYO}} &
  \multirow{4}{*}{Sentinel-2} &
  X &
  3D-ConvLSTM &
  0.48 {\tiny $\pm$ \textcolor{black!50}{0.29}} &
  2.68 {\tiny $\pm$ \textcolor{black!50}{0.33}} &
  0.54 {\tiny $\pm$ \textcolor{black!50}{0.18}} &
  1.04 {\tiny $\pm$ \textcolor{black!50}{0.16}} &
  0.67 {\tiny $\pm$ \textcolor{black!50}{0.58}} &
  1.38 {\tiny $\pm$ \textcolor{black!50}{0.32}} &
  0.15 {\tiny $\pm$ \textcolor{black!50}{0.11}} &
  2.67 {\tiny $\pm$ \textcolor{black!50}{0.90}} &
  0.19 {\tiny $\pm$ \textcolor{black!50}{0.11}} &
  1.08 {\tiny $\pm$ \textcolor{black!50}{0.17}} &
  0.06 {\tiny $\pm$ \textcolor{black!50}{0.19}} &
  1.52 {\tiny $\pm$ \textcolor{black!50}{0.45}} &
  0.21 {\tiny $\pm$ \textcolor{black!50}{0.15}} &
  1.49 {\tiny $\pm$ \textcolor{black!50}{0.31}} &
  0.15 {\tiny $\pm$ \textcolor{black!50}{0.27}} &
  2.71 {\tiny $\pm$ \textcolor{black!50}{0.79}} &
  0.34 {\tiny $\pm$ \textcolor{black!50}{0.09}} &
  1.28 {\tiny $\pm$ \textcolor{black!50}{0.12}} \\
\hline
\end{tabular}

    } 
    \label{tab:table_all_loyo_subfield}
\end{table*}
\begin{table*}[!tb]
    \centering
    \caption{
    Results for the RMSE (t/ha) ($\downarrow$) and the $R^2$-score ($\uparrow$) for different models and datasets for the Leave-One-Region-Out (LORO) cross-validation at the field level. ARG = Argentina, BRA = Brazil, GER = Germany, URG = Uruguay. C = corn, R = rapeseed, S = soybean, W = wheat.
    }
    \resizebox{.95\linewidth}{!}{
\begin{tabular}{ccclllllllllllllllllll}
\hline
\multicolumn{4}{c}{\multirow{2}{*}{\textbf{Evaluation}}} &
  \multicolumn{18}{c}{\textbf{Field-Level}} \\ \cline{5-22} 
\multicolumn{4}{c}{} &
  \multicolumn{2}{c}{\textbf{ARG-C}} &
  \multicolumn{2}{c}{\textbf{ARG-S}} &
  \multicolumn{2}{c}{\textbf{ARG-W}} &
  \multicolumn{2}{c}{\textbf{BRA-C}} &
  \multicolumn{2}{c}{\textbf{BRA-S}} &
  \multicolumn{2}{c}{\textbf{BRA-W}} &
  \multicolumn{2}{c}{\textbf{GER-R}} &
  \multicolumn{2}{c}{\textbf{GER-W}} &
  \multicolumn{2}{c}{\textbf{URG-S}} \\ \hline
\textbf{Splitting} &
  \textbf{Modalities} &
  \textbf{Fusion Method} &
  \multicolumn{1}{c}{\textbf{Model}} &
  $R^2$ &
  RMSE &
  $R^2$ &
  RMSE &
  $R^2$ &
  RMSE &
  $R^2$ &
  RMSE &
  $R^2$ &
  RMSE &
  $R^2$ &
  RMSE &
  $R^2$ &
  RMSE &
  $R^2$ &
  RMSE &
  $R^2$ &
  RMSE \\ \hline
\multirow{10}{*}{\textbf{LORO}} &
  \multirow{4}{*}{Sentinel-2} &
  X &
  3D-ConvLSTM &
  0.68 {\tiny $\pm$ \textcolor{black!50}{1.89}} &
  1.59 {\tiny $\pm$ \textcolor{black!50}{0.48}} &
  0.69 {\tiny $\pm$ \textcolor{black!50}{0.12}} &
  0.67 {\tiny $\pm$ \textcolor{black!50}{0.12}} &
  0.80 {\tiny $\pm$ \textcolor{black!50}{0.21}} &
  0.95 {\tiny $\pm$ \textcolor{black!50}{0.24}} &
  0.55 {\tiny $\pm$ \textcolor{black!50}{0.74}} &
  1.19 {\tiny $\pm$ \textcolor{black!50}{0.36}} &
  0.37 {\tiny $\pm$ \textcolor{black!50}{1.63}} &
  0.55 {\tiny $\pm$ \textcolor{black!50}{0.18}} &
  0.47 {\tiny $\pm$ \textcolor{black!50}{0.30}} &
  0.58 {\tiny $\pm$ \textcolor{black!50}{0.24}} &
  0.25 {\tiny $\pm$ \textcolor{black!50}{0.53}} &
  1.16 {\tiny $\pm$ \textcolor{black!50}{0.31}} &
  0.14 {\tiny $\pm$ \textcolor{black!50}{1.05}} &
  1.76 {\tiny $\pm$ \textcolor{black!50}{0.75}} &
  0.68 {\tiny $\pm$ \textcolor{black!50}{0.52}} &
  0.61 {\tiny $\pm$ \textcolor{black!50}{0.30}} \\
\hline
\end{tabular}

    } 
    \label{tab:table_all_loro_field_leve}
\end{table*}

\begin{table*}[!tb]
    \centering
    \caption{
    Results for the RMSE (t/ha) ($\downarrow$) and the $R^2$-score ($\uparrow$) for different models and datasets for the Leave-One-Region-Out (LORO) cross-validation at the subfield (pixel)-level. ARG = Argentina, BRA = Brazil, GER = Germany, URG = Uruguay. C = corn, R = rapeseed, S = soybean, W = wheat.
    }
    \resizebox{.95\linewidth}{!}{
\begin{tabular}{ccclllllllllllllllllll}
\hline
\multicolumn{4}{c}{\multirow{2}{*}{\textbf{Evaluation}}} &
  \multicolumn{18}{c}{\textbf{Subfield (Pixel)-Level}} \\ \cline{5-22} 
\multicolumn{4}{c}{} &
  \multicolumn{2}{c}{\textbf{ARG-C}} &
  \multicolumn{2}{c}{\textbf{ARG-S}} &
  \multicolumn{2}{c}{\textbf{ARG-W}} &
  \multicolumn{2}{c}{\textbf{BRA-C}} &
  \multicolumn{2}{c}{\textbf{BRA-S}} &
  \multicolumn{2}{c}{\textbf{BRA-W}} &
  \multicolumn{2}{c}{\textbf{GER-R}} &
  \multicolumn{2}{c}{\textbf{GER-W}} &
  \multicolumn{2}{c}{\textbf{URG-S}} \\ \hline
\textbf{Splitting} &
  \textbf{Modalities} &
  \textbf{Fusion Method} &
  \multicolumn{1}{c}{\textbf{Model}} &
  $R^2$ &
  RMSE &
  $R^2$ &
  RMSE &
  $R^2$ &
  RMSE &
  $R^2$ &
  RMSE &
  $R^2$ &
  RMSE &
  $R^2$ &
  RMSE &
  $R^2$ &
  RMSE &
  $R^2$ &
  RMSE &
  $R^2$ &
  RMSE \\ \hline
\multirow{10}{*}{\textbf{LORO}} &
  \multirow{4}{*}{Sentinel-2} &
  X &
  3D-ConvLSTM &
  0.54 {\tiny $\pm$ \textcolor{black!50}{0.38}} &
  2.52 {\tiny $\pm$ \textcolor{black!50}{0.33}} &
  0.57 {\tiny $\pm$ \textcolor{black!50}{0.11}} &
  1.01 {\tiny $\pm$ \textcolor{black!50}{0.10}} &
  0.69 {\tiny $\pm$ \textcolor{black!50}{0.24}} &
  1.33 {\tiny $\pm$ \textcolor{black!50}{0.23}} &
  0.33 {\tiny $\pm$ \textcolor{black!50}{0.14}} &
  2.35 {\tiny $\pm$ \textcolor{black!50}{0.37}} &
  0.19 {\tiny $\pm$ \textcolor{black!50}{0.22}} &
  1.08 {\tiny $\pm$ \textcolor{black!50}{0.33}} &
  0.16 {\tiny $\pm$ \textcolor{black!50}{0.07}} &
  1.44 {\tiny $\pm$ \textcolor{black!50}{0.31}} &
  0.12 {\tiny $\pm$ \textcolor{black!50}{0.29}} &
  1.57 {\tiny $\pm$ \textcolor{black!50}{0.30}} &
  0.07 {\tiny $\pm$ \textcolor{black!50}{0.43}} &
  2.84 {\tiny $\pm$ \textcolor{black!50}{0.69}} &
  0.36 {\tiny $\pm$ \textcolor{black!50}{0.19}} &
  1.26 {\tiny $\pm$ \textcolor{black!50}{0.29}} \\
\hline
\end{tabular}

    } 
    \label{tab:table_all_loro_subfield}
\end{table*}

\clearpage
\section{Dataset Datasheet}\label{sec:datasheet}
In the following, we will refer to the authors and their institutions in the paper as the “publishing institution.” 

\subsection{Motivation}
The motivation to create the dataset was to advance large-scale crop yield prediction. Data creation was funded by industry and public stakeholders. 

\subsection{Dataset Creation}
The dataset can be used for crop yield prediction as described in the main paper. The dataset can be used for further research or other directions, such as crop classification, time series analysis, or land cover classification, provided that data privacy and the data license are complied with. 

\subsection{Data Collection Process}
All the yield data was collected in collaboration with local farmers, data providers, and industry companies. Each data provider had to sign a data-sharing agreement with the publishing institution. Data providers were compensated for data collection and data sharing on an area basis.  \\
The \gls{eo} data was acquired by the publishing institution from only publicly available data providers under the licenses of those providers. \\
The dataset does not contain any other data from other publicly available data sources or libraries and is entirely self-contained. 
The data contains sensitive information about the geographic locations of the local data providers. Each data point is georeferenced. We provide georeferences to facilitate downstream research, including additional data acquisition activities beyond the current dataset. \\
The dataset may have errors and inconsistencies arising during the data collection process. This may include errors in the yield data collection and preprocessing processes. The dataset may include local biases to individual regions, crop types, and years. Moreover, the data may include data imbalances that could bias downstream models.
We provide detailed metadata to help researchers account for these limitations.

\subsection{Data Preprocessing}

The data preprocessing was described in detail in Sec. \ref{sec:preprocessing}. Statistics to calculate the target scaled yield (dry yield) are given in Tab. \ref{tab:yield_threshold}.

\subsection{Data Distribution}
The publishing institution owns the data in its entirety and will grant access under a custom license. 

\subsection{Data Maintenance}
The dataset is entirely self-contained. 
The publishing institution will maintain the data entirely on its internal servers.  
The publishing institution will maintain the data access portal and provide sufficient information to contact it. The data will be continuously maintained and updated. Updates will be documented on the publishing institution's webpage. We welcome community contributions that improve the data quality. 

The dataset is available at \url{https://yieldsat.github.io/}

\subsection{Legal and Ethical Considerations}
The data is owned entirely by the publishing institution. Access will be granted under the described license and data privacy regulation. 
The data does not contain any information about human subjects or any other confidential information. However, sensitive information is provided in the georeferences of the data. 

This dataset is intended for academic research in a non-commercial setting. There are no expectations for this dataset to produce models that can generalize to real-world data, and the data usage is aimed for academic research, for example: to build new yield estimation models, evaluate yield estimation models, and study the yield estimation process itself.


\end{document}